\documentclass{article}

\usepackage{microtype}
\usepackage{graphicx}
\usepackage{booktabs} % for professional tables
\usepackage{kotex}
\usepackage{amsmath,amssymb}
\usepackage[dvipsnames]{xcolor}
\usepackage{etoolbox}
\usepackage{breqn}
\usepackage{algorithm} 
\usepackage{algpseudocode}
\usepackage{setspace}
\usepackage{siunitx}
\usepackage{bm}
\usepackage{lipsum} % for dummy text
\usepackage{enumitem}
\usepackage{chngcntr}

\usepackage{caption}
\usepackage{comment}
\usepackage{makecell}

\usepackage{xfrac}
\usepackage{multirow}
\usepackage{mathtools}

\usepackage{subcaption}
\usepackage{caption}

\usepackage{hyperref}

\usepackage[accepted]{icml2021}

\icmltitlerunning{Submission and Formatting Instructions for ICML 2021}
\begin{document}

\twocolumn[
\icmltitle{Mitigating Memorization in Sample Selection for Learning with Noisy Labels}



\icmlsetsymbol{equal}{*}

\begin{icmlauthorlist}
\icmlauthor{Kyeongbo Kong}{equal,POSTECH}
\icmlauthor{Junggi Lee}{equal,POSTECH}
\icmlauthor{Youngchul Kwak}{POSTECH}
\icmlauthor{Young-Rae Cho}{SAMSUNG}
\icmlauthor{Seong-Eun Kim}{SeoulTech}
\icmlauthor{Woo-Jin Song}{POSTECH}
\end{icmlauthorlist}

\icmlaffiliation{POSTECH}{Department of Electrical Engineering, Pohang University of Science and Technology, Pohang 37673, South Korea}
\icmlaffiliation{SAMSUNG}{Samsung institude of advanced technology, suwon-si, 16678, South Korea}
\icmlaffiliation{SeoulTech}{Department of Applied Artificial Intelligence, Seoul National University of Science and Technology (SeoulTech), Seoul 01811, South Korea}

\icmlcorrespondingauthor{Woo-Jin Song}{wjsong@postech.ac.kr}

\icmlkeywords{Machine Learning, ICML}

\vskip 0.3in
]

\printAffiliationsAndNotice{\icmlEqualContribution} % otherwise use the standard text.



%\title{How to Reduce Memorization in Learning with Noisy Labels}
%%\title{Mitigation of Memorization in Sample Selection for Robust Learning of Deep Neural Networks with Noisy Labels}


\begin{abstract}

%Deep learning is known to be vulnerable to noisy labels, hence learning methods with noisy labels have been actively studied. Sample selection techniques, which train networks with only correct-labeled data, have attracted a great attention. 

Because deep learning is vulnerable to noisy labels, \textit{sample selection techniques}, which train networks with only clean labeled data, have attracted a great attention. However, if the labels are dominantly corrupted by few classes, these noisy samples are called dominant-noisy-labeled samples, the network also learns dominant-noisy-labeled samples rapidly via content-aware optimization. In this study, we propose a compelling criteria to penalize dominant-noisy-labeled samples intensively through class-wise penalty labels. By averaging prediction confidences for the each observed label, we obtain suitable penalty labels that have high values if the labels are largely corrupted by some classes. Experiments were performed using benchmarks (CIFAR-10, CIFAR-100, Tiny-ImageNet) and real-world datasets (ANIMAL-10N, Clothing1M) to evaluate the proposed criteria in various scenarios with different noise rates. Using the proposed sample selection, the learning process of the network becomes significantly robust to noisy labels compared to existing methods in several noise types.
\end{abstract}




\section{Introduction}
\label{Introduction}

%Recently, deep neural networks (DNNs) have shown exceptional performances in several computer vision applications  \citep{wang2015feedforward, qian2018simple, ijjina2016hybrid, chevtchenko2018convolutional,  he2017mask, kong2019multitask, ilg2017flownet, lee2020blocknet, chen2016cnntracker, fernando2018soft+}. 

For training deep neural networks (DNNs) using a supervised learning method, the labeled data are predominantly obtained from web queries \cite{liu2011noise}, crowdsourcing \cite{welinder2010multidimensional, han2018robust, yan2014learning}, e-commerce \cite{xiao2015learning} and social-network \cite{cha2012social}. However, several noisy labeled samples (In this study, we omit ``labeled'' for brevity) that are annotated with erroneous or noisy labels are present in the data. Zhang \cite{zhang2016understanding} confirmed that a DNN is significantly vulnerable to noisy samples because it can adequately learn (memorize) their patterns. To robustly train a network regardless of noisy samples, learning with noisy labels has been studied actively. These studies can be divided into two categories based on the technique employed: loss correction and sample selection.

In the loss correction category, the loss (or label) of a mini-batch is modified using a noise transition matrix \cite{goldberger2016training, patrini2017making, han2018masking, hendrycks2018using} or prediction confidences of samples \cite{reed2014training, chang2017active, tanaka2018joint, ren2018learning, wang2019symmetric}. However, if the number of classes or noise rate is large, the prediction confidences and noise transition matrix can be inaccurate \cite{han2018co, song2019selfie}, leading to false correction and error accumulation in the training stage.

\begin{figure*}[t]
\centering
\subcaptionbox{\label{fig1:subfig1}}{
    \includegraphics[width=10.5cm]{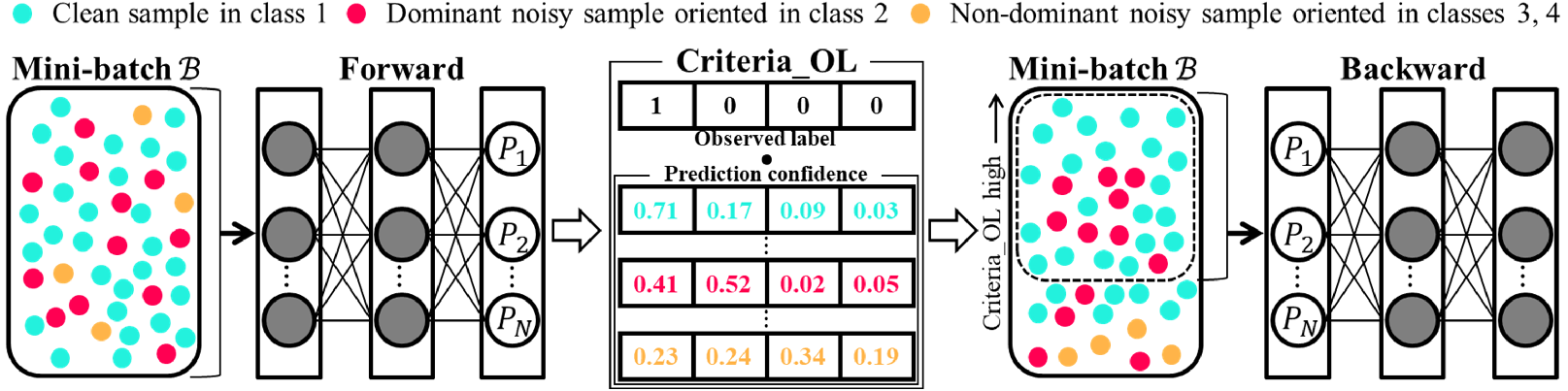}
}
\subcaptionbox{\label{fig1:subfig2}}{
    \includegraphics[width=13.2cm]{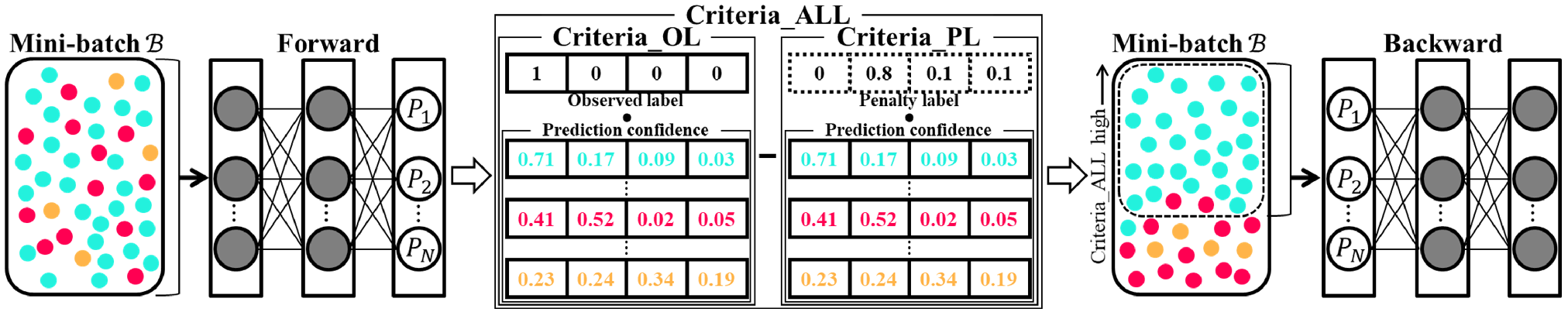}
}

\caption{Comparison between (a) conventional and (b) proposed algorithms based on robust sample selection. In mini-batch $\mathcal{B}$, for clear visualization, we only represent samples which have a single observed label among various observed labels. In Criteria with Observed Label (Criteria\_OL) and Criteria with Penalty Label (Criteria\_PL), inner product of prediction confidence of each sample and the label is performed. Furthermore, $P_{i}$ represents the prediction confidence in the $i$th class. In the sample selection stage, the conventional algorithm using Criteria\_OL selects lots of dominant noisy samples as well as clean samples. In contrast, the proposed algorithm using Criteria\_ALL, which subtracts a proposed Criteria\_PL from Criteria\_OL, rarely selects dominant noisy samples because dominant noisy samples can be penalized appropriately through Criteria\_PL with proposed penalty label. Thereafter, both algorithms updates network parameters using the selected samples with their observed labels.}
\label{fig1_framework}
\end{figure*}

Recently, to overcome this limitation, many studies have been based on the sample selection method. In the sample selection category, only clean samples, whose labels are accurate, are selected from the mini-batch and thereafter used to train the network \cite{han2018co, malach2017decoupling, wang2019co, yu2019does, shen2019learning,song2019selfie, kong2019recycling}`. They are based on ``content-aware optimization'' (also called ``memorization effect''), wherein it initially learns simple patterns shared by multiple training samples across the data regardless of their labels \cite{arpit2017closer}. Using this characteristic, sample selection based algorithms (Fig. \ref{fig1_framework}a) assume that the DNN initially learns clean samples more rapidly than noisy ones. Therefore, the algorithms first select samples which have high scores (low losses) using observed labels that include noisy labels, then, train the DNN using the selected samples.

However, the performances of the sample selection algorithms vary depending on types of noise. If the labels are dominantly corrupted by some classes, noisy samples generally have shared patterns and can also be learned rapidly by content-aware optimization (we call these noisy samples \textbf{dominant noisy samples}). Thus, dominant noisy samples have high score with observed labels, thus selection algorithms can not exclude these samples. In Fig. \ref{fig1_framework}(a), many dominant noisy samples are selected for training DNN because dominant noisy samples have higher scores than the other noisy samples and some clean samples. They cause severe memorization of DNN, so it is important to penalize dominant noisy samples in sample selection. 

In this study, we introduce Criteria\_ALL, which subtracts new \textbf{Criteria with the proposed Penalty Label (Criteria\_PL)} from the conventional Criteria\_OL with the observed label for penalizing dominant noisy samples (Fig. \ref{fig1_framework}b). In the proposed Criteria\_PL with penalty label, dominant noisy samples have higher scores than the non-dominant ones and clean samples. Therefore, Criteria\_ALL is effective for penalizing dominant noisy samples (Fig. \ref{fig1_framework}b; red samples). To estimate the penalty label, we use the average-prediction confidence of samples that belong to identical observed label after normalizing the result (Section \ref{Criteria ALL}; Fig. \ref{fig2_penalty_label}). In summary, the main contributions of this study are the following:
\begin{itemize}
\item We observe that dominant noisy samples cause memorization (reduce generalization) of DNN through empirical evidence.
\item We propose a new Criteria\_PL that have high criteria scores for dominant noisy samples, and can penalize dominant noisy samples by subtracting Criteria\_PL from the conventional Criteria\_OL (Criteria\_ALL).
\end{itemize}

\section{Main methodology: penalizing dominant noisy samples in sample selection}
\label{Main methodology}
In this section, we describe our sample selection method to achieve robust learning in detail.
\subsection{Sample selection with conventional Criteria\_OL}
\label{Conventional sample selection}

In the $K$-class classification problem, let ${\mathcal{D}=\left\{\left({\mathbf{x}}_{i}, {\hat{\mathbf{y}}}_{i} \right) \right\}}_{i=1}^{n}$ be training dataset, where ${\mathbf{x}}_{i}$ is the $i$th sample and ${\hat{\mathbf{y}}}_{i}$ is the $i$th observed label that includes corrupted label and is in the label space $\left\{{\mathbf{e}}^{k}: k\in \left\{1,2,\cdots,K \right\} \right\}$, i.e., ${\mathbf{e}}^{k}$:$k$th one-hot vector such as $\left[0,\cdots,0,1,0,\cdots,0 \right]$. The objective is to train the DNN parameter $\boldsymbol{\theta}$ through the maximum likelihood estimation ${\mathrm{argmax}}_{\boldsymbol{\theta}}\sum_{i=1}^{n}P\left({\hat{\mathbf{y}}}_{i}|{\mathbf{x}}_{i};\boldsymbol{\theta} \right)$. Herein, the prediction confidence in $j$th class, which indicates the probability of being assigned to $j$th class (probability for the label $\mathbf{e}^{j}$), is obtained by applying a softmax function to the DNN $f$:
\begin{eqnarray}
P\left(\mathbf{y}=\mathbf{e}^{j}|{\mathbf{x}}_{i};\boldsymbol{\theta} \right)=\frac{\exp\left\{{f\left({\mathbf{x}}_{i} \right)}_{j} \right\}}{\sum_{l=1}^{K}\exp\left\{{f\left({\mathbf{x}}_{i} \right)}_{l} \right\}}, \forall j=1,\cdots,K,\nonumber
\label{eq1}
\end{eqnarray}
where ${f\left(\cdot \right)}_{j}$  is $j$th component of $f\left(\cdot \right)$. 
The loss function is generally defined by the cross-entropy scheme with the observed label ${\mathbf{\hat{\mathbf{y}}}_{i}}$ and prediction  $P\left(\mathbf{y}=\mathbf{e}^{j}|{\mathbf{x}}_{i};\boldsymbol{\theta} \right)$:
\begin{eqnarray}
\mathcal{L}\left({\mathbf{x}}_{i}, {\hat{\mathbf{y}}}_{i};\boldsymbol{\theta} \right)=-\log P\left(\mathbf{y}={\hat{\mathbf{y}}}_{i}|{\mathbf{x}}_{i}; \boldsymbol{\theta} \right).
\label{eq2}
\end{eqnarray}
DNN is trained through a stochastic optimization method, which updates the parameter $\boldsymbol{\theta}$ using the mini-batch procedure:
\begin{eqnarray}
{\boldsymbol{\theta}}^{t+1}={\boldsymbol{\theta}}^{t}-\eta \nabla\left(\frac{1}{\left|\mathcal{B} \right|}\sum_{\left({\mathbf{x}}_{i},{\hat{\mathbf{y}}}_{i} \right)\in \mathcal{B}}\mathcal{L}\left({\mathbf{x}}_{i}, {\hat{\mathbf{y}}}_{i};{\boldsymbol{\theta}}^{t} \right) \right),
\label{eq3}
\end{eqnarray}
where ${\boldsymbol{\theta}}^{t}$ is a parameter in the $t$th iteration, $\eta$ is the learning rate, and $\mathcal{B}$ is the mini-batch fetched from $\mathcal{D}$.

Recently, several algorithms based on sample selection method \cite{han2018co,jiang2017mentornet,wang2019co,yu2019does} have been developed to robustly train the DNN for noisy labels. These methods assume that clean samples are trained more easily compared with the noisy samples, because DNN is trained by content-aware optimization wherein it initially learns simple patterns shared among multiple samples \cite{arpit2017closer}. Therefore, in a mini-batch manner, sample selection methods select top $R\%$ of sorted samples in descending order relative to probability for the observed label (Fig. \ref{fig1_framework}a). Then the network parameter is updated using only selected samples as
\begin{eqnarray}
{\boldsymbol{\theta}}^{t+1}={\boldsymbol{\theta}}^{t}-\eta \nabla\left(\frac{1}{\left|{\mathcal{S}} \right|}\sum_{\left({\mathbf{x}}_{i},{\hat{\mathbf{y}}}_{i} \right)\in {\mathcal{S}}}\mathcal{L}\left({\mathbf{x}}_{i}, {\hat{\mathbf{y}}}_{i};{\boldsymbol{\theta}}^{t} \right) \right),
\label{eq4}
\end{eqnarray}
where
\begin{align}
{\mathcal{S}}
&={\mathrm{SELECT}}_{\left({\mathbf{x}}_{i},{\hat{\mathbf{y}}}_{i} \right)\in {\mathcal{B}}}^{\mathrm{TOP}\,R\%}\sum_{j=1}^{K}{\hat{\mathbf{y}}}_{i}\left(j \right)P\left(\mathbf{y}={\mathbf{e}}^{j}|{\mathbf{x}}_{i}; {\boldsymbol{\theta}}^{t} \right),
\label{eq5}
\end{align}
where ${\mathrm{TOP}\,R\%}$ means top $R\%$ of sorted samples in descending order, ${\hat{\mathbf{y}}}_{i}\left(\cdot \right)$ is an element of one-hot vector ${\hat{\mathbf{y}}}_{i}$, $R\%=100-\varepsilon$, and  $\varepsilon$ is the percentage of noise label. It is based on selection criteria, Criteria\_OL, where ``sample should have high prediction confidence for observed label'' as follows: 
\begin{eqnarray}
{\mathrm{{Criteria}\_{OL}}\left({\mathbf{x}}_{i},{\hat{\mathbf{y}}}_{i};{\boldsymbol{\theta}}^{t} \right)=\sum_{j=1}^{K}{\hat{\mathbf{y}}}_{i}\left(j \right)P\left(\mathbf{y}={\mathbf{e}}^{j}|{\mathbf{x}}_{i}; {\boldsymbol{\theta}}^{t} \right)}.
\label{eq6}\nonumber
\end{eqnarray}

However, if the labels are largely corrupted by some classes, these samples have high probability of having several shared patterns among themselves, like clean labels data. Therefore, dominant noisy samples can also be included in selected samples, causing memorization (reducing generalization) of DNN (empirical evidence is provided in Section \ref{Why neccessary}).

\subsection{Criteria\_ALL: combination of proposed Criteria\_PL and Criteria\_OL}
\label{Criteria ALL}

To penalize dominant noisy samples, we propose the criteria, Criteria\_PL, which exploits proposed penalty label. In Criteria\_PL, dominant noisy samples have higher values than the non-dominants. This is because the penalty label is set to have high values in the classes where dominant noisy samples originate; moreover, (dominant) noisy samples generally have high prediction confidences in their true class. Therefore, to penalize dominant noisy samples, we introduce a selection criteria (Criteria\_ALL) by subtracting Criteria\_PL ``sample should have high prediction confidence for proposed penalty label $\tilde{\mathbf{y}}$'' from Criteria\_OL as follows:
\begin{eqnarray}
\mathrm{{Criteria}\_{ALL}}=\mathrm{{Criteria}\_{OL}}- \lambda \: \mathrm{{Criteria}\_{PL}},
\label{eq7}
\end{eqnarray}
where
\begin{eqnarray}
{\mathrm{{Criteria}\_{OL}}\left({\mathbf{x}}_{i},{\hat{\mathbf{y}}}_{i};{\boldsymbol{\theta}}^{t} \right)=\sum_{j=1}^{K}{\hat{\mathbf{y}}}_{i}\left(j \right)P\left(\mathbf{y}={\mathbf{e}}^{j}|{\mathbf{x}}_{i}; {\boldsymbol{\theta}}^{t} \right)},\nonumber\\
\mathrm{{Criteria}\_{PL}}\left({\mathbf{x}}_{i}, {\tilde{\mathbf{y}}}_{i}; {\boldsymbol{\theta}}^{t} \right)=\sum_{j=1}^{K}{\tilde{\mathbf{y}}}_{i}\left(j \right) P\left(\mathbf{y}=\mathbf{e}^{j}|{\mathbf{x}}_{i};{\boldsymbol{\theta}}^{t} \right),
\label{eq8}\nonumber
\end{eqnarray}
where $\lambda$ is hyperparameter (user design parameter) and  ${\tilde{\mathbf{y}}}_{i}\left(\cdot \right)$ is an element of $i$th penalty label. The effectiveness of Criteria\_PL is described in detail with empirical evidence in Section \ref{Why neccessary}.

\begin{figure}[t]
\centering
\includegraphics[width=0.5\textwidth]{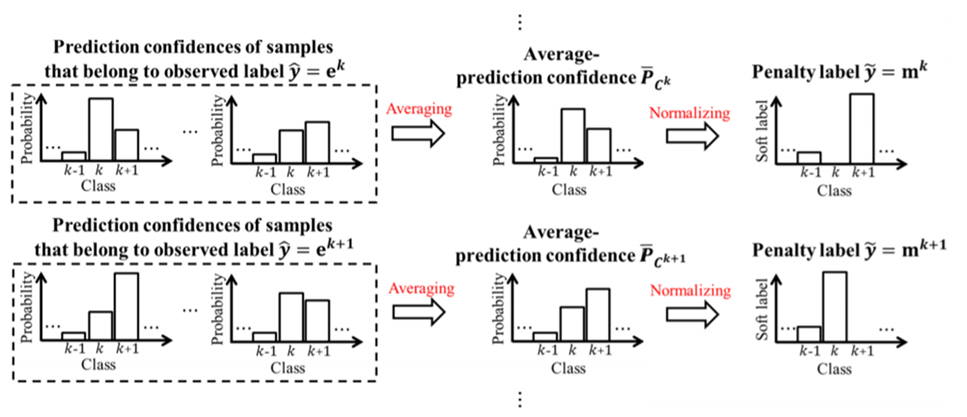}
\caption{Process of acquiring penalty label ${\tilde{\mathbf{y}}}$: this process is conducted for each observed label ${\hat{\mathbf{y}}}=\mathbf{e}^{k}$ in the same manner.}
\label{fig2_penalty_label}
\end{figure}

To penalize dominant noisy samples through the proposed Criteria\_ALL, it is necessary to estimate the penalty label properly. The process to obtain class-wise penalty label ${\tilde{\mathbf{y}}}$ is shown in Fig. \ref{fig2_penalty_label}. First, we average the prediction confidences among samples that have same observed label ${C}^{k}=\left\{\left({\mathbf{x}}_{i}, {\hat{\mathbf{y}}}_{i} \right): \hat{\mathbf{y}}_{i}=\mathbf{e}^{k} \right\}$ to measure the extent of dominance of the noisy samples. This method is reasonable because the percentage of noisy samples that are dominant is high; the dominant noisy samples have high prediction confidence in their true class (this will be shown in Section \ref{Why neccessary}). Finally, to ensure the same scale with observed labels, the compensated average-prediction confidence is normalized except the class that the observed label $\hat{\mathbf{y}}=\mathbf{e}^{k}$ indicates (labeled class $k$). This is because samples in labeled class $k$ are not needed to be penalized. 
%It is expressed mathematically as follows. 

The penalty label $\tilde{\mathbf{y}}_{i}\in\left\{{\mathbf{m}}^{k}: k\in \left\{1,2,\cdots,K \right\} \right\}$ is defined by
\begin{eqnarray}
{\mathbf{m}}^{k}\left(j \right)=
\begin{cases}
\dfrac{1}{A} {\bar{P}}_{{C}^{k}}\left(\mathbf{y}=\mathbf{e}^{j}|{\boldsymbol{\theta}}^{t} \right), & \text{ if } \; j\neq k  \\ 
0, & \text{ if } \; j= k 
\end{cases},
\label{eq9}
\end{eqnarray}
where
\begin{eqnarray}
A=\sum_{l \neq k}{ \bar{P}}_{{C}^{k}}\left(\mathbf{y}=\mathbf{e}^{l}|{\boldsymbol{\theta}}^{t} \right),
\label{eq10}
\end{eqnarray}
\begin{eqnarray}
{{\bar{P}}_{{C}^{k}}\left(\mathbf{y}=\mathbf{e}^{j}|{\boldsymbol{\theta}}^{t} \right)=\sum_{\left({\mathbf{x}}_{i}, {\hat{\mathbf{y}}}_{i} \right)\in {C}^{k}}{P}\left(\mathbf{y}=\mathbf{e}^{j}|{\mathbf{x}}_{i};\boldsymbol{\theta}^{t} \right)},
\label{eq11}
\end{eqnarray}
where ${C}^{k}$ is $\left\{\left({\mathbf{x}}_{i}, {\hat{\mathbf{y}}}_{i} \right): \hat{\mathbf{y}}_{i}=\mathbf{e}^{k} \right\}$. 

%The weight $\mathbf{w}$, used for compensating the learning speed per class, increases if the average-prediction confidence of the entire training data is lower than the prior distribution $\boldsymbol{q}$ as follows
%\begin{eqnarray}
%\mathbf{w}\left(j \right) = \frac{\boldsymbol{q}\left(j \right)}{\frac{1}{\left|\mathcal{D} \right|}\sum_{\left({\mathbf{x}}_{i}, {\hat{\mathbf{y}}}_{i} \right)\in \mathcal{D}}{P}\left(\mathbf{y}=\mathbf{e}^{j}|{\mathbf{x}}_{i};\boldsymbol{\theta}^{t} \right)},
%\label{eq12}
%\end{eqnarray}
%where $\boldsymbol{q}\left(\cdot \right)$ is an element of the prior distribution, indicating the class distribution of the entire training data \cite{tanaka2018joint}. Further details of weight $\mathbf{w}$ is presented in Section \ref{Compensating_weight}.

\begin{algorithm}[t]
\caption{$\textrm{Sample selection with Criteria\_ALL.}$}
\begin{algorithmic}[1]
\For{$t=1:num\_epochs$} 
	\For{$l=1:num\_iterations$}
		\State \textbf{Predict} mini-batch ${\mathcal{B}}$;
		\State {\textbf{Stack} prediction confidences $P$;}
		\State \textbf{Select} samples ${\mathcal{S}}_{Prop}$ from mini-batch ${\mathcal{B}}$ within \State \quad\quad\,\,\, top $R\%$ using $\mathrm{Criteria\_ALL}$;
		\State \textbf{Update} network using selected samples ${\mathcal{S}}_{Prop}$;
	\EndFor
	\State {\textbf{Update} penalty label $\tilde {\mathbf{y}}^{t}$;}
\EndFor
\end{algorithmic}
\label{algo1}
\end{algorithm}

Finally, the network is trained with samples selected using $\text{{Criteria}\_{ALL}}$ by 
\begin{align}
{{\mathcal{S}}}_{Prop}&={\mathrm{SELECT}}_{\left({\mathbf{x}}_{i},{\hat{\mathbf{y}}}_{i} \right)\in {\mathcal{B}}}^{\mathrm{TOP}\,R\%}\mathrm{{Criteria}\_{ALL}}\left({\mathbf{x}}_{i}, {\hat{\mathbf{y}}}_{i}, {\tilde{\mathbf{y}}}_{i}; {\boldsymbol{\theta}}^{t} \right).
\label{eq13}
\end{align}
Algorithm \ref{algo1} describes the overall procedure of sample selection with Criteria\_ALL. The penalty label is updated at the end of each epoch (in step 9 of Algorithm \ref{algo1}).

\begin{figure}[!h]
\centering
\subcaptionbox{\label{fig3:subfig1}}{%
    \includegraphics[width=3.8cm]{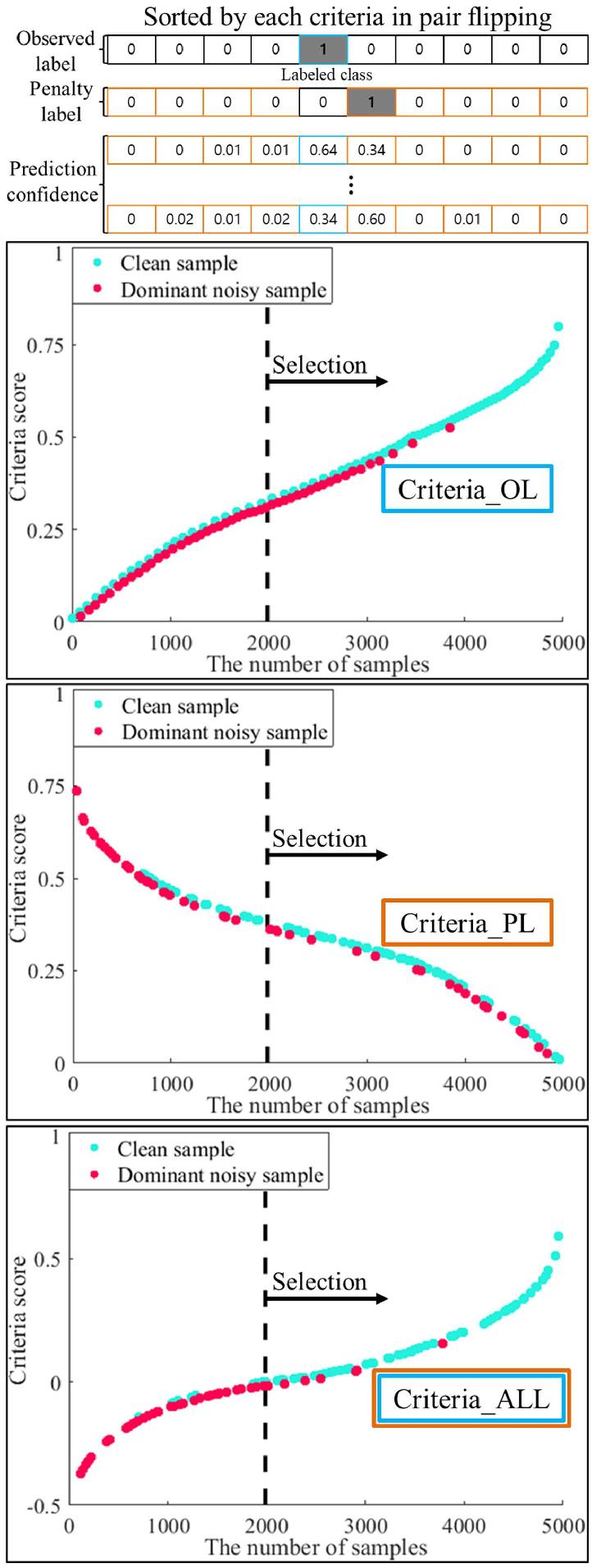}
} \hspace*{-0.83em}
\subcaptionbox{\label{fig3:subfig2} \hspace*{-1.0em}}{  
    \includegraphics[width=3.8cm]{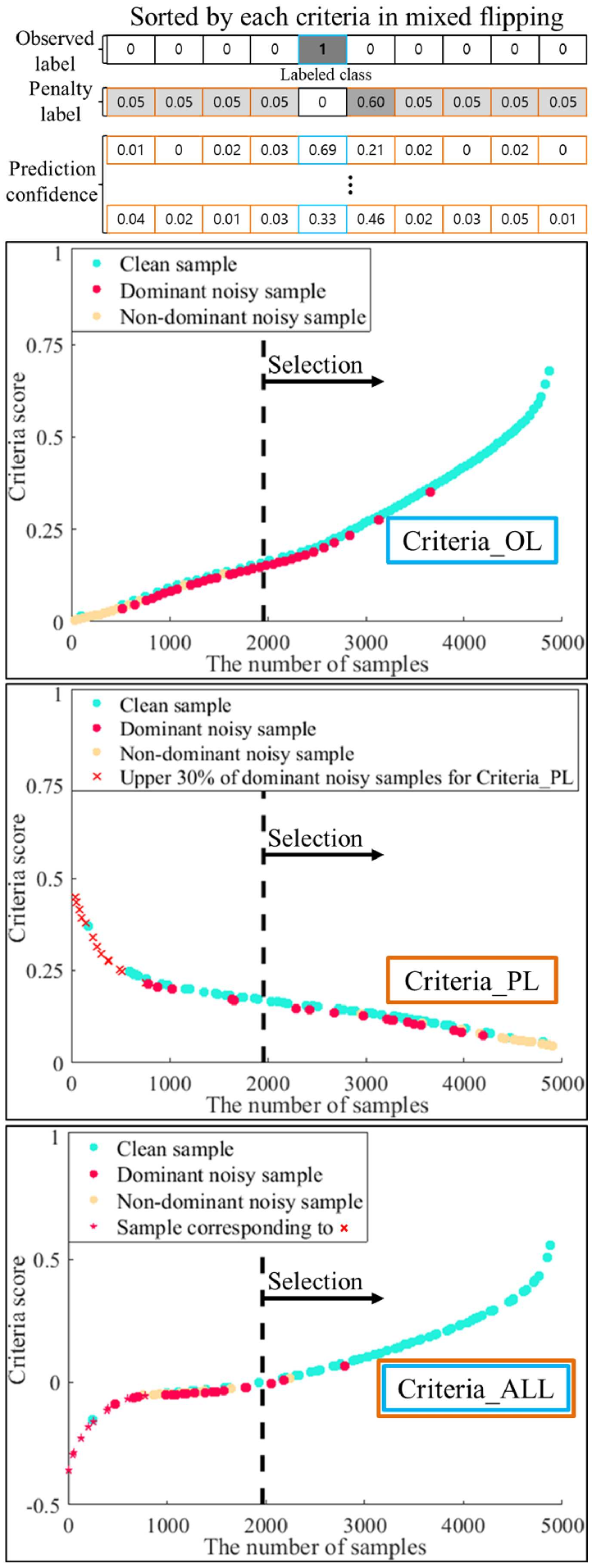}
}\hspace*{-0.6em}
\caption{When the sample selection is performed with pair and mixed flipping noise, the results are represented by sorted samples in each criteria. The samples on the right-hand side of the black dashed line are selected. For clear visualization, we only show samples that have same observed label $\hat{\mathbf{y}}=\mathbf{e}^{5}$ on CIFAR-10 in initial training. Furthermore, we conduct down-sampling by interval 40, then depict these ones sorting for each criteria. The Criteria\_ALL and Criteria\_OL are sorted in ascending order, and the Criteria\_PL is sorted in descending order. (a) Pair flipping \cite{han2018co}: the true labels of class 5 are corrupted by the labels of class 6 with the noise rate of 40\%. (b) Mixed flipping: the true labels of class 5 are corrupted by the labels of class 6 with the noise rate of 24\% and the other classes at a same noise rate of 2\% (total 40\%).}
\label{fig3_concept}
\end{figure}

\section{Deep understanding for Criteria\_ALL}
\label{Deep understanding}
In this section, we mainly describe the effectiveness of Criteria\_ALL based on empirical evidence on CIFAR-10 dataset \cite{krizhevsky2009learning}.

\subsection{Why Criteria\_OL and Criteria\_PL are necessary?}
\label{Why neccessary}
%In \cite{arpit2017closer}, the DNN is trained by content-aware optimization where it initially learns simple patterns shared among multiple samples. In sample selection based on noisy label fields, several algorithms utilize this characteristic to distinguish noisy samples from training data. They assume that the clean samples are trained more rapidly than noisy samples, i.e., clean samples exhibit high prediction confidence for observed labels in initial training, because clean samples have considerable shared patterns with themselves. Therefore, conventional sample selection methods select the top $R\%$ of samples using Criteria\_OL and train DNN to become robust to noisy labels. However, if labels are corrupted by pair noise types, some noisy samples get included in selected samples (Fig. \ref{fig3_concept} Criteria\_OL; red circles on the right-hand side of the black dashed line), leading to memorization (reducing generalization) of DNN.

We explain the validity of Criteria\_ALL with respect to the following two points.
\begin{enumerate}[label={\arabic*.}]
\item Which noisy samples should be penalized in sample selection with Criteria\_OL?
\item How can Criteria\_PL penalize noisy samples using the proposed penalty label?
\end{enumerate}

\begin{figure}[t]
\centering
\subcaptionbox{\label{fig7:subfig1} \hspace*{-2.1em}}{
    \includegraphics[width=2.5cm]{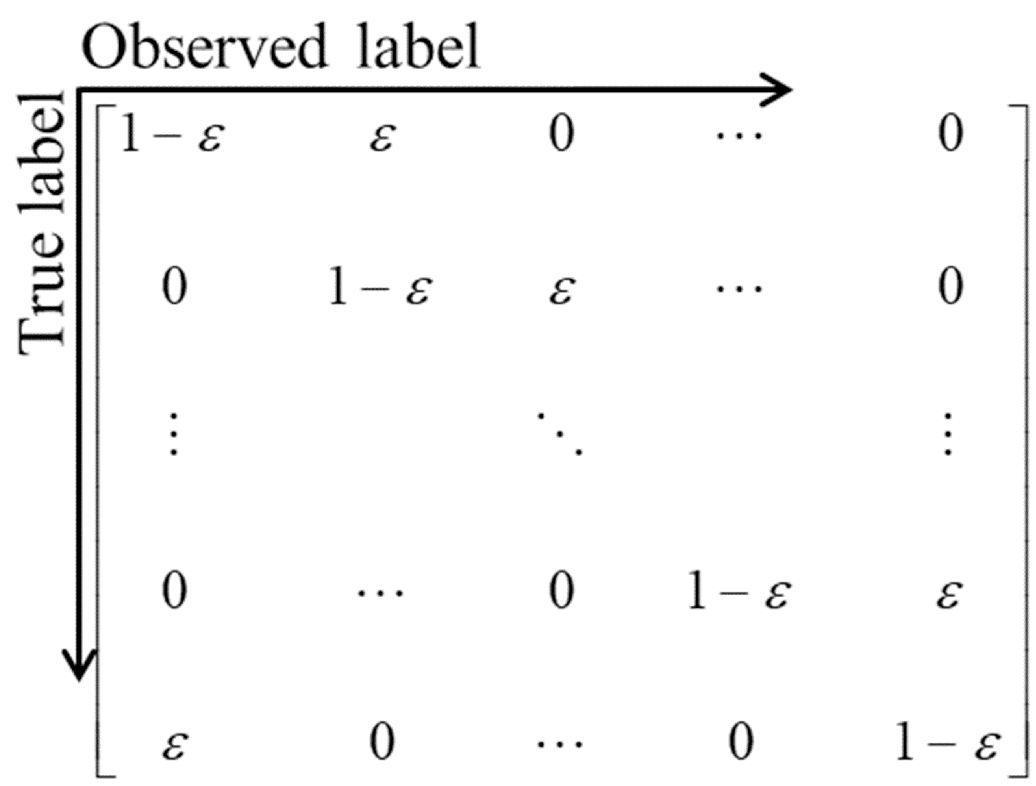}
}
\subcaptionbox{\label{fig7:subfig2} \hspace*{-2.2em}}{
    \includegraphics[width=2.5cm]{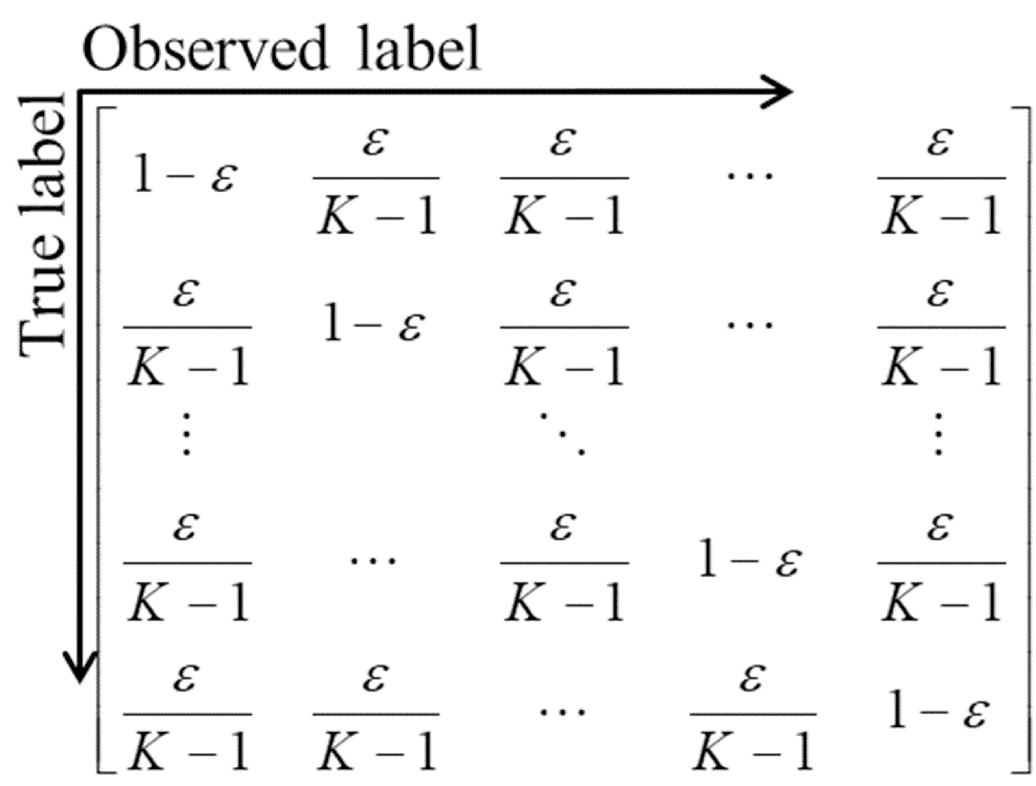}
}
\subcaptionbox{\label{fig7:subfig3} \hspace*{-2.3em}}{
    \includegraphics[width=2.5cm]{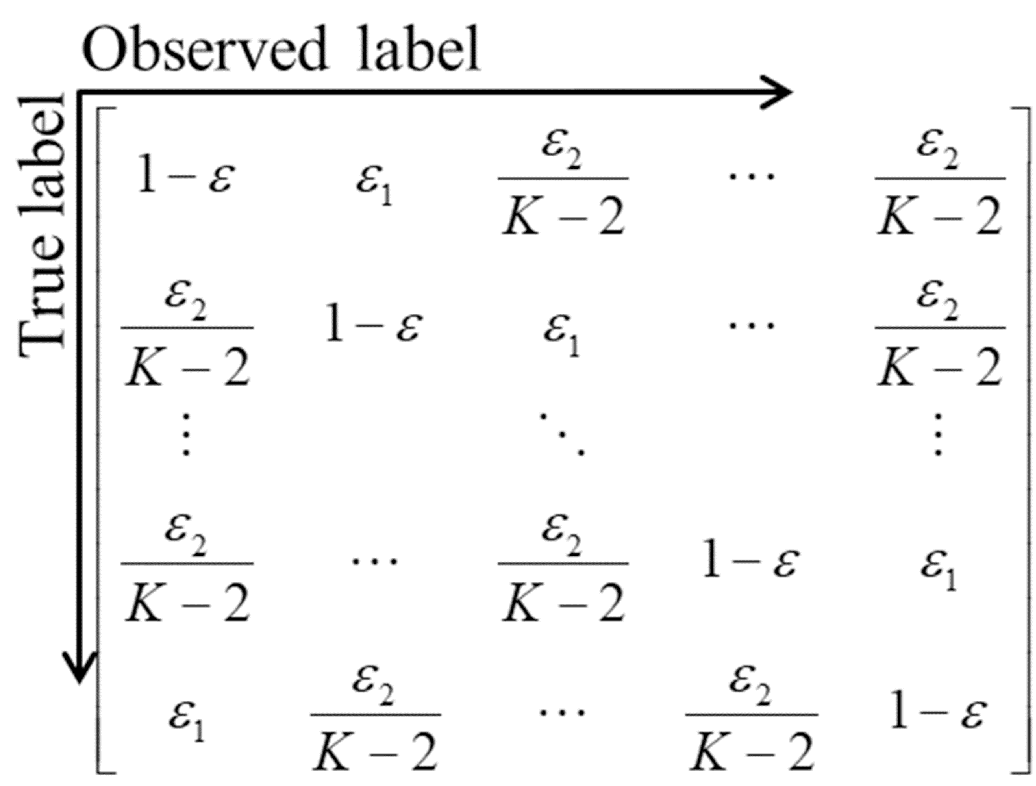}
}
\caption{{Several noise types; (a) Pair flipping; (b) Symmetry flipping; (c) Mixed flipping ($\varepsilon = \varepsilon_{1} +  \varepsilon_{2}$).}}
\label{fig7_noise_trainsition_matrix}
\end{figure}

First, when the labels are significantly corrupted by certain classes (pair $>$ mixed $>$ symmetry in Fig. \ref{fig7_noise_trainsition_matrix}), dominant noisy samples generally have shared patterns with themselves, and thus can also be learned rapidly by content-aware optimization. In pair flipping, because noisy samples are corrupted by a single class (all noisy samples are dominant), several noisy samples are included in the selected samples (Fig. \ref{fig3_concept}a top panel; red circles on the right-hand side of the black dashed line). In mixed flipping, noisy samples originate in several classes and can be dominant or non-dominant depending on their proportion. Because dominant noisy samples have considerable shared patterns compared to the non-dominants, the dominants are trained more rapidly than the non-dominants by content-aware optimization; consequently, some noisy samples among dominant noisy samples are included in selected samples (Fig. \ref{fig3_concept}b top panel; red solid circles on the right-hand side of the black dashed line). Therefore, \textit{dominant noisy samples cause memorization (reduce generalization) of DNN, and it is important to penalize them in sample selection.}

%In mixed flipping, noisy samples originate in several classes and can be dominant or non-dominant depending on their proportion. Because dominant noisy samples have considerable shared patterns compared to the non-dominants, the dominants are trained more rapidly than the non-dominants by content-aware optimization; consequently, some noisy samples among dominant noisy samples are included in selected samples (Fig. \ref{fig3_concept}b top panel; red solid circles on the right-hand side of the black dashed line). In symmetry flipping, because there are no dominant noisy samples, few noisy samples are included in the selected samples (Fig. \ref{fig3_concept}c top panel; yellow solid circles on the right-hand side of the black dashed line).

Thereafter, we focus on how Criteria\_PL can penalize dominant noisy samples using penalty labels. For general noisy label distributions (e.g., mixed flipping case), because dominant noisy samples cause significant memorization than non-dominant ones, we propose a novel class-wise penalty label that reflects extent of dominance for penalizing noisy samples differently. When the penalty label and prediction confidence of each sample are multiplied by inner product, then the dominant noisy samples will have higher values than that of the non-dominants. This is because (dominant) noisy samples generally have high prediction confidence in their true class, and the penalty label is set to have relatively higher value in the class, which dominant noisy samples originates, than the other classes. Therefore, we can distinguish dominant noisy sample using the proposed criteria, Criteria\_PL, defined by inner product of penalty label and prediction confidence of each sample. In Fig. \ref{fig3:subfig2}, because penalty label is set to have a relatively high value (e.g., sixth value in penalty label: 0.60) in class 6 where dominant noisy samples originate, and prediction confidences of dominant noisy samples also have the highest value (e.g., sixth value in last prediction confidence: 0.46) in class 6 (its true class), most dominant noisy samples have higher criteria scores than clean and non-dominant samples in Criteria\_PL (middle panel). Finally, we subtract the Criteria\_PL from Criteria\_OL (defined as Criteria\_ALL) for penalizing dominant noisy samples largely. Using Criteria\_ALL, penalized dominant and non-dominant noisy samples have smaller criteria scores compared to most clean samples and have similar criteria scores with each other (Fig. \ref{fig3:subfig2} bottom panel). Lastly, in symmetry flipping case, Criteria\_PL does not affect the selection of clean samples because of no dominant noisy samples (proofed in Appendix \ref{Special cases Symmetry}).

\section{Experiments}
\label{Experiment results}
We evaluate the effect of the proposed Criteria\_ALL with regard to two perspectives. First, we evaluate the performance using three benchmark datasets by corrupting clean labels using three different type of noisy label with various noise rates. Second, we evaluate the performance using two real-world datasets (Appendix \ref{real_dataset}).
Three independent trials were performed and average and standard error are used to represent the performance of the proposed, and conventional algorithms. Dataset are described in (Appendix \ref{Dataset}).

\begin{figure*}[t]
\centering
\subcaptionbox{\label{fig8:subfig1} \hspace*{-3.3em}}{
    \includegraphics[width=4.2cm]{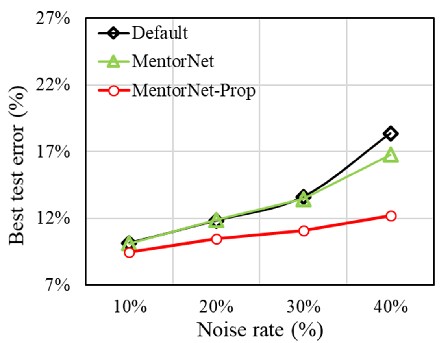}
}
\subcaptionbox{\label{fig8:subfig2} \hspace*{-3.6em}}{
    \includegraphics[width=4.2cm]{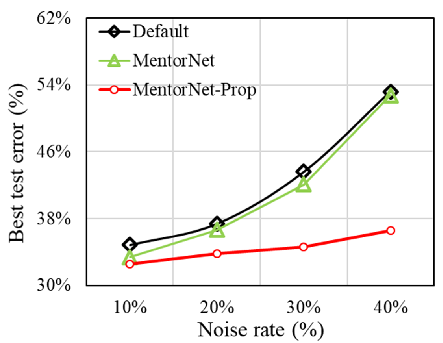}
}
\subcaptionbox{\label{fig8:subfig3} \hspace*{-3.9em}}{
    \includegraphics[width=4.2cm]{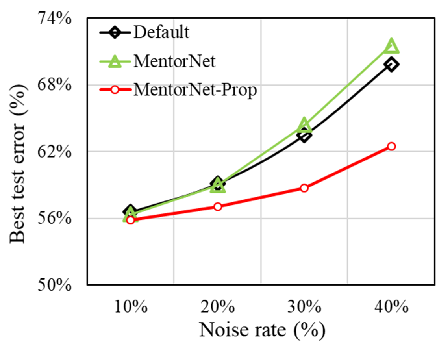}
}
\caption{Best test error (\%) based on various noise rates with pair flipping; (a) CIFAR-10; (b) CIFAR-100; (c) Tiny-ImageNet.}
\label{fig8_bench_pair}
\end{figure*}
\subsection{Algorithms}
\label{System_setup}
We compared Criteria\_ALL with several algorithms in the loss correction and sample selection categories. $\textit{Default}$ (baseline) was network that was trained without processing for noisy labels. $\textit{Active Bias}$ \cite{chang2017active} (loss correction) trained DNN by assigning large weights to uncertain samples that exhibit high variance. $\textit{MentorNet}$ \cite{jiang2017mentornet} (sample selection) used Criteria\_OL to select clean samples and trains DNN using only selected samples. $\textit{Co-teaching}$ \cite{han2018co} (sample selection) used two networks and exchanged information of sample selection to enhance the difference. $\textit{Co-teaching+}$ (sample selection) applied Co-teaching to the samples with different prediction confidence. $\textit{SL}$ \cite{wang2019symmetric} (loss correction) combined reverse cross entropy term with cross entropy loss to enhance cross entropy's learning on hard classes and tolerance to noisy labels. $\textit{SELFIE}$ \cite{song2019selfie} (hybrid) corrected loss of samples with high precision while conducting the conventional sample selection. $\textit{MentorNet--Prop}$ is a basic sample selection method that uses Criteria\_ALL rather than Criteria\_OL. Because Criteria\_ALL can be applied flexibly to other algorithms, we also conducted experiments using \textit{SL--Prop} and \textit{SELFIE--Prop} by combining Criteria\_ALL with SL and SELFIE (Appendix \ref{Collaboration}).

\begin{table*}[t]
\centering
\caption{Best test error (\%) with different noise rates for each noise type}
\resizebox{\textwidth}{!}{%
\begin{tabular}{c|ccccccccc}
\hline
                 & Default & Active Bias & Co-teaching & Co-teaching+ & SL & SELFIE & \textbf{MentorNet--Prop} & \textbf{SL--Prop} & \textbf{SELFIE--Prop} \\ \Xhline{3\arrayrulewidth}
Sample selection & $\times$      & $\times$           & $\bigcirc$        & $\bigcirc$             & $\times$          & $\bigcirc$ & $\bigcirc$      & $\bigcirc$     & $\bigcirc$          \\ 
Loss correction  & $\times$      & $\bigcirc$           & $\times$        & $\times$             & $\bigcirc$          & $\bigcirc$ & $\times$      & $\bigcirc$     & $\bigcirc$   \\ \hline     
\end{tabular}%
}
\begin{tabular}{c}
\\ 
\end{tabular}
\resizebox{\textwidth}{!}{%
\setlength{\tabcolsep}{4pt} % Default value: 6pt    col
\renewcommand{\arraystretch}{0.8} % Default value: 1  row
\begin{tabular}{c|c|cccc|cccc|c}

\hline
\multirow{3}{*}{Dataset}                                                   & \multirow{3}{*}{Method} & \multicolumn{4}{c|}{Pair flipping}                     & \multicolumn{4}{c|}{Symmetry flipping}                 & Mixed      \\ \cline{3-11} 
                                                                           &                         & \multicolumn{4}{c|}{Noise rate $\varepsilon$}               & \multicolumn{4}{c|}{Noise rate $\varepsilon$}               & Noise rate $\varepsilon$\\
                                                                           &                         & 10\%      & 20\%      & 30\%      & 40\%      & 0\%       & 20\%      & 40\%      & 60\%      & 40\%       \\ \Xhline{3\arrayrulewidth}
\multirow{10}{*}{\begin{tabular}[c]{@{}c@{}}CIFAR-\\ 10\end{tabular}}      & Default                 & 10.2$\pm$0.24 & 11.8$\pm$0.03 & 13.6$\pm$0.07 & 18.4$\pm$0.35 & 8.6$\pm$0.07  & 13.2$\pm$0.08 & 17.4$\pm$0.25 & 25.7$\pm$0.55 & 16.0$\pm$0.16  \\
																  & Active Bias             & 9.5$\pm$0.08  & 10.8$\pm$0.13 & 12.1$\pm$0.09 & 17.6$\pm$0.33 & 8.5$\pm$0.09  & 12.2$\pm$0.16 & 15.7$\pm$0.39 & 24.2$\pm$0.21 & 14.4$\pm$0.23  \\
                                                                           & Co-teaching             & 10.0$\pm$0.09 & 11.3$\pm$0.13 & 12.7$\pm$0.01 & 15.3$\pm$0.30 & 8.8$\pm$0.05  & 11.0$\pm$0.01 & 14.1$\pm$0.18 & \textbf{19.4$\pm$0.12} & 14.4$\pm$0.36  \\
                                                                           & Co-teaching+            & 12.4$\pm$0.13 & 14.0$\pm$0.21 & 15.8$\pm$0.22 & 23.1$\pm$2.17 & 11.2$\pm$0.25 & 13.7$\pm$0.17 & 17.2$\pm$0.30 & 33.5$\pm$1.90 & 17.9$\pm$0.49  \\
                                                                           %& MentorNet               & 10.1$\pm$0.14 & 11.9$\pm$0.19 & 13.4$\pm$0.07 & 16.8$\pm$0.55 & 8.8$\pm$0.10  & 11.4$\pm$0.24 & 14.4$\pm$0.02 & 21.0$\pm$0.54 & 15.2$\pm$0.27  \\
                                                                           
                                                                           & SL                      & 10.2$\pm$0.12 & 11.2$\pm$0.03 & 13.7$\pm$0.20 & 18.3$\pm$0.46 & 8.5$\pm$0.04  & 12.6$\pm$0.17 & 17.0$\pm$0.32 & 25.6$\pm$0.16 & 16.2$\pm$0.15  \\
                                                                           & SELFIE                  & 9.5$\pm$0.10  & 10.6$\pm$0.08 & 11.5$\pm$0.05 & 13.3$\pm$0.08 & 8.7$\pm$0.09  & 10.9$\pm$0.01 & \textbf{13.5$\pm$0.04} & 20.5$\pm$0.23 & 13.3$\pm$0.17  \\
                                                                           & \textbf{MentorNet--Prop} & 9.5$\pm$0.05  & 10.5$\pm$0.04 & 11.1$\pm$0.04 & 12.2$\pm$0.12 & 8.6$\pm$0.05  & 11.1$\pm$0.11 & 14.4$\pm$0.26 & 20.5$\pm$0.11 & 13.4$\pm$0.11  \\
                                                                           & \textbf{SL--Prop}        & 9.3$\pm$0.13  & 10.2$\pm$0.07 & \textbf{10.8$\pm$0.18} & \textbf{11.7$\pm$0.13} & 8.6$\pm$0.13  & 11.0$\pm$0.07 & 13.7$\pm$0.03 & 19.9$\pm$0.08 & 13.2$\pm$0.23  \\                                                                    
                                                                           & \textbf{SELFIE--Prop}    & \textbf{8.9$\pm$0.13}  & \textbf{9.7$\pm$0.07}  & 10.8$\pm$0.12 & 11.7$\pm$0.11 & \textbf{8.4$\pm$0.09}  & \textbf{10.5$\pm$0.02} & 13.7$\pm$0.10 & 20.4$\pm$0.10 & \textbf{13.0$\pm$0.11}  \\ \hline
\multirow{10}{*}{\begin{tabular}[c]{@{}c@{}}CIFAR-\\ 100\end{tabular}}     & Default                 & 34.9$\pm$0.29 & 37.4$\pm$0.13 & 43.6$\pm$0.49 & 53.2$\pm$0.46 & 31.7$\pm$0.18 & 37.5$\pm$0.18 & 43.5$\pm$0.21 & 53.8$\pm$0.08 & 47.6$\pm$0.71  \\
																  & Active Bias             & 34.1$\pm$0.32 & 36.6$\pm$0.16 & 40.9$\pm$0.27 & 50.8$\pm$0.08 & 31.1$\pm$0.31 & 36.6$\pm$0.28 & 42.1$\pm$0.29 & 52.9$\pm$0.19 & 44.8$\pm$0.14  \\
                                                                           & Co-teaching             & 33.2$\pm$0.20 & 36.2$\pm$0.13 & 41.4$\pm$0.15 & 51.8$\pm$0.75 & 31.9$\pm$0.16 & 34.3$\pm$0.11 & 37.5$\pm$0.01 & 44.3$\pm$0.57 & 45.7$\pm$0.42  \\
                                                                           & Co-teaching+            & 35.6$\pm$0.17 & 37.2$\pm$0.17 & 40.3$\pm$0.18 & 47.3$\pm$0.20 & 33.8$\pm$0.19 & 36.2$\pm$0.15 & 40.6$\pm$0.21 & 49.1$\pm$0.31 & 43.7$\pm$0.18  \\
                                                                           %& MentorNet               & 33.4$\pm$0.41 & 36.7$\pm$0.32 & 42.1$\pm$0.53 & 52.8$\pm$0.69 & 31.7$\pm$0.29 & 34.5$\pm$0.20 & 38.2$\pm$0.05 & 46.1$\pm$0.80 & 46.7$\pm$0.54  \\
                                                                           & SL                      & 34.1$\pm$0.14 & 36.5$\pm$0.08 & 42.0$\pm$0.42 & 52.3$\pm$0.59 & 31.1$\pm$0.05 & 36.5$\pm$0.30 & 42.7$\pm$0.23 & 52.4$\pm$0.53 & 45.7$\pm$0.23  \\
                                                                           & SELFIE                  & 32.7$\pm$0.09 & 34.1$\pm$0.12 & 35.8$\pm$0.18 & 41.3$\pm$0.15 & 31.7$\pm$0.05 & 33.8$\pm$0.08 & 37.1$\pm$0.02 & 44.6$\pm$0.13 & 38.5$\pm$0.16  \\
                                                                           & \textbf{MentorNet--Prop} & 32.5$\pm$0.04 & 33.8$\pm$0.19 & 34.6$\pm$0.20 & 36.6$\pm$0.11 & 31.3$\pm$0.16 & 34.6$\pm$0.10 & 38.3$\pm$0.11 & 45.4$\pm$0.34 & 37.8$\pm$0.23  \\
                                                                           & \textbf{SL--Prop}        & \textbf{32.4$\pm$0.11} & 33.3$\pm$0.11 & 34.4$\pm$0.09 & 36.2$\pm$0.40 & 31.1$\pm$0.23 & 34.4$\pm$0.07 & 38.0$\pm$0.11 & 46.0$\pm$0.63 & 37.2$\pm$0.42  \\
                                                                           
                                                                           & \textbf{SELFIE--Prop}    & 32.6$\pm$0.14 & \textbf{33.1$\pm$0.03} & \textbf{34.1$\pm$0.16} & \textbf{35.3$\pm$0.19} & \textbf{30.9$\pm$0.17} & \textbf{33.7$\pm$0.18} & \textbf{36.6$\pm$0.15} & \textbf{42.6$\pm$0.05} & \textbf{36.0$\pm$0.29}  \\ \hline
\multirow{10}{*}{\begin{tabular}[c]{@{}c@{}}Tiny-\\ ImageNet\end{tabular}} & Default                 & 56.6$\pm$0.07 & 59.1$\pm$0.45 & 63.5$\pm$0.18 & 69.9$\pm$0.28 & 54.5$\pm$0.13 & 59.2$\pm$0.35 & 64.6$\pm$0.21 & 72.9$\pm$0.62 & 66.4$\pm$0.20  \\
                                                                           & Active Bias             & 56.6$\pm$0.05 & 58.9$\pm$0.12 & 62.1$\pm$0.29 & 68.2$\pm$0.06 & 55.0$\pm$0.23 & 58.7$\pm$0.11 & 64.1$\pm$0.35 & 72.2$\pm$0.14 & 65.0$\pm$0.01  \\
                                                                           & Co-teaching             & 56.3$\pm$0.13 & 59.0$\pm$0.15 & 63.7$\pm$0.38 & 71.5$\pm$0.19 & 54.9$\pm$0.07 & 56.9$\pm$0.29 & 60.9$\pm$0.13 & 65.7$\pm$0.08 & 66.8$\pm$0.12  \\
                                                                           & Co-teaching+            & 58.3$\pm$0.11 & 60.0$\pm$0.15 & 62.9$\pm$0.13 & 69.3$\pm$0.46 & 56.9$\pm$0.16 & 59.0$\pm$0.09 & 63.1$\pm$0.13 & 71.6$\pm$0.21 & 65.7$\pm$0.28  \\
                                                                           %& MentorNet               & 56.4$\pm$0.20 & 59.0$\pm$0.31 & 64.4$\pm$0.20 & 71.6$\pm$0.13 & 54.9$\pm$0.14 & 57.7$\pm$0.35 & 60.8$\pm$0.14 & 67.2$\pm$0.28 & 66.9$\pm$0.24  \\
                                                                           & SL                      & 56.0$\pm$0.22 & 58.4$\pm$0.35 & 61.9$\pm$0.14 & 67.5$\pm$0.26 & 54.1$\pm$0.10 & 58.4$\pm$0.12 & 64.4$\pm$0.14 & 72.3$\pm$0.28 & 64.5$\pm$0.25  \\
                                                                           & SELFIE                  & 56.1$\pm$0.19 & 57.9$\pm$0.11 & 59.7$\pm$0.06 & 64.7$\pm$0.08 & 54.7$\pm$0.19 & \textbf{56.4$\pm$0.22} & 60.3$\pm$0.08 & 69.4$\pm$0.19 & 63.3$\pm$0.18  \\
                                                                           & \textbf{MentorNet--Prop} & 55.8$\pm$0.10 & 57.1$\pm$0.12 & 58.7$\pm$0.14 & 62.5$\pm$0.22 & 54.9$\pm$0.06 & 57.7$\pm$0.04 & 61.0$\pm$0.10 & 67.0$\pm$0.29 & 60.9$\pm$0.17  \\

                                                                           & \textbf{SL--Prop}        & \textbf{54.9$\pm$0.07} & \textbf{56.8$\pm$0.23} & 58.0$\pm$0.29 & \textbf{60.1$\pm$0.26} & \textbf{54.0$\pm$0.04} & 57.5$\pm$0.21 & 60.9$\pm$0.11 & 66.9$\pm$0.26 & 60.7$\pm$0.30  \\
                                                                           
                                                                           & \textbf{SELFIE--Prop}    & 55.5$\pm$0.04 & 56.8$\pm$0.13 & \textbf{57.6$\pm$0.20} & 60.5$\pm$0.21 & 54.1$\pm$0.10 & 57.1$\pm$0.12 & \textbf{59.7$\pm$0.21} & \textbf{64.0$\pm$0.01} & \textbf{60.1$\pm$0.17}  \\ \hline
\end{tabular}%
}

\label{table2}
\end{table*}

\subsection{Experimental results on benchmark datasets}
\label{benchmark_datasets}

\begin{table}[!t]
\centering
\small
\caption{Best test error (\%) in mixed flipping}
\label{table1}
\scalebox{0.9}{
\begin{tabular}{c|ccc}
\hline
Algorithm       & CIFAR-10  & CIFAR-100 & Tiny-ImageNet \\ \Xhline{3\arrayrulewidth}
Default         & 16.0$\pm$0.16 & 47.6$\pm$0.71 & 66.4$\pm$0.20     \\ \hline
MentorNet       & 15.2$\pm$0.27 & 46.7$\pm$0.54 & 66.9$\pm$0.24     \\ \hline
\textbf{MentorNet--Prop} & \textbf{13.4$\pm$0.11} & \textbf{37.8$\pm$0.23} & \textbf{60.9$\pm$0.17}     \\ \hline
\end{tabular}}
\end{table}
%for pair, symmetry, and mixed flipping
To examine the effect of the proposed Criteria\_ALL in the sample selection category, we first compared the test errors of Default (without Criteria), MentorNet (with Criteria\_OL), and MentorNet--Prop (with Criteria\_ALL). In pair flipping, MentorNet--Prop presented the lowest test errors at all noise rates and for all datasets (Fig. \ref{fig8_bench_pair}), particularly when the noise rate was high. For mixed flipping, we conducted experiments with noise rate 40\% (Table \ref{table1}). Similar to pair flipping, MentorNet--Prop recorded the lowest test error in three datasets.

Thereafter, we compared MentorNet--Prop, SL--Prop, and SELFIE--Prop with conventional algorithms in the loss correction and sample selection categories with various noise rates. In Table \ref{table2} pair flipping, MentorNet--Prop displayed the overall lower test error compared to other conventional algorithms. Furthermore, SL--Prop and SELFIE--Prop achieved lowest test error. When the proposed Criteria\_ALL is applied to SL, the test errors obtained were $0.9$--$16.1\%$ lower than those of SL. This means selecting sample using Criteria\_ALL is also effective in loss correction category. SELFIE--Prop achieved test errors $0.1$--$6.0\%$ lower than those of SELFIE. 
%In addition, SELFIE--Prop exhibited a high performance even after learning once (Table \ref{table3}; pair flipping). That is, improving the distinguishing ability in sample selection is helpful in hybrid category. 
In conclusion, when labels are corrupted as pair flipping, proposed Criteria\_ALL is powerful. %Additionally, Co-teaching+ recorded the worse performance than expected. This is because two networks are rapidly consensus and several samples are excluded too early in complex network \cite{song2019prestopping}.

The performances of the algorithms at various noise rates are also compared for the symmetry case in Table \ref{table2}. MentorNet--Prop achieved lower or similar test errors as the conventional algorithms. Particularly, when the proposed Criteria\_ALL is applied to SL, the test errors were $0.9$--$6.4\%$ lower than in SL. Experimental results show that Criteria\_ALL is also effective in the loss correction category in symmetry flipping case. %Meanwhile, SELFIE--Prop had lower or similar test errors as SELFIE because of SELFIE setting . Because sample selection with Criteria\_ALL can distinguish clean samples well compared to Criteria\_OL in pair flipping, for SELFIE--Prop, we can use a tight hyperparameter, which is suitable in symmetry flipping.   
%(uncertainty threshold $= 0.005$ and history length $= 25$)
For mixed flipping case, MentorNet--Prop achieved lower or similar test errors to the conventional algorithms. Particularly, SELFIE--Prop had the lowest test errors. %Consequently, it is verified that the proposed Criteria is effective in general noise type. 

\section{Conclusion}
\label{Conclusion}
To address learning with noisy labels, we introduced a novel criteria, Criteria\_ALL that penalizing dominant noisy samples effectively by subtracting the proposed Criteria\_PL from the conventional Criteria\_OL. Because the penalty label reflects the extent of dominance of noisy samples, dominant noisy samples have higher criteria scores compared to non-dominant noisy samples and clean samples in Criteria\_PL. Therefore, Criteria\_ALL is effective for penalizing dominant noisy samples by subtracting Criteria\_PL. 
%Particularly, it is more effective for extremely memorized noisy samples because Criteria\_ALL can penalize these samples intensively. Furthermore, to accurately estimate the penalty label, we propose additional techniques that compensate learning speed per class and update based on a temporal ensembling. 
From our experiments, sample selection using the Criteria\_ALL was effective for several benchmark and real-world datasets. 
%In further investigations, the authors intend to perform both sample selection and loss correction using the penalty label.

\section{Acknowledgements}
\label{Acknowledgements}
This research was supported in part by Basic Science Research Program through the National Research Foundation of Korea (NRF) funded by the Ministry of Education (2020R1A6A3A01098940, 2021R1I1A1A01051225), the MSIT (Ministry of Science and ICT), Korea, under the ICT Consilience Creative program (IITP-2019-2011-1-00783) supervised by the IITP (Institute for Information \& communications Technology Planning \& Evaluation), National Research Foundation of Korea (NRF) funded by the Korean government (2020R1C1C1011857, 2020M3C1B8081320), and LG Display under LGD-POSTECH Research Program.

\bibliography{example_paper}
\bibliographystyle{icml2021}





\newpage

\appendix

% \counterwithin*{equation}{part}

\setcounter{equation}{0}

%\addcontentsline{toc}{section}{First Section}
\section{Additional details for improving class-wise penalty label and compatibility of Criteria\_ALL}
\label{Additional details}
With Criteria\_ALL, dominant noisy samples are effectively penalized in sample selection. In this section, we describe additional consideration when Criteria\_ALL is applied to the noisy label fields. 

\subsection{Improvement of class-wise penalty label}
\label{Class-wise penalty label}

In Sections \ref{subsub_update}, the effectiveness of the update method is described in detail.

\subsubsection{Updating penalty label using temporal ensembling}
\label{subsub_update}

\begin{algorithm}[!h]
\caption{$\textrm{Update penalty label using prediction}$ \newline $\textrm{confidences at the end of an epoch}$}
\begin{algorithmic}[1]
\For{$t=1:num\_epochs$} 
	\For{$l=1:num\_iterations$}
		\State \textbf{Predict} mini-batch ${\mathcal{B}}$;
%		\State {\textbf{Stack} prediction confidences $P$;}
		\State \textbf{Select} samples ${\mathcal{S}}_{Prop}$ from mini-batch ${\mathcal{B}}$ within \State \quad\quad\,\,\, top $R\%$ using $\mathrm{Criteria\_ALL}$;
		\State \textbf{Update} network using selected samples ${\mathcal{S}}_{Prop}$;
	\EndFor
	\State \textcolor{red}{\textbf{Re-predict} all samples $\mathcal{D}$;}
	\State \textcolor{red}{\textbf{Update} penalty label $\tilde {\mathbf{y}}^{t}$ through $\textit{prediction confidences}$;}
\EndFor
\end{algorithmic}
\label{algo2}
\end{algorithm}
\begin{algorithm}[!h]
\caption{$\textrm{Update penalty label using prediction}$ \newline $\textrm{confidences for every iteration}$}
\begin{algorithmic}[1]
\For{$t=1:num\_epochs$} 
	\For{$l=1:num\_iterations$}
		\State \textbf{Predict} mini-batch ${\mathcal{B}}$;
		\State \textcolor{red}{{\textbf{Stack} prediction confidences $P$;}}
		\State \textbf{Select} samples ${\mathcal{S}}_{Prop}$ from mini-batch ${\mathcal{B}}$ within \State \quad\quad\,\,\, top $R\%$ using $\mathrm{Criteria\_ALL}$;
		\State \textbf{Update} network using selected samples ${\mathcal{S}}_{Prop}$;
	\EndFor
	\State \textcolor{red}{\textbf{Update} penalty label $\tilde {\mathbf{y}}^{t}$ through $\textit{stacked prediction confidences}$;}
\EndFor
\end{algorithmic}
\label{algo3}
\end{algorithm}

When a DNN is robustly trained (considering noisy label), accurate average-prediction confidence and penalty label estimation are likely to occur. Therefore, penalty label should be updated for each epoch. Two types of methods are available for estimating the penalty label during the training stage.

In single model method (Algorithm \ref{algo2}), new prediction confidences for all the training data are estimated at the end of each epoch. This results in a new penalty label as well, which is used in the next epoch. However, as the parameters of the DNN are updated iteratively by a stochastic optimization method, it can approach bad local minima with respect to generalization \cite{huang2017snapshot}. When the DNN falls into the local minima at the end of an epoch in the initial training, the penalty label is inaccurately estimated. This can adversely affect the learning of the DNN in the next epoch.

In our ensemble method (Algorithm \ref{algo3}; we re-describe Algorithm \ref{algo1} as Algorithm \ref{algo3} for clear comparison), the prediction confidences for the mini-batch are stored per iteration, and the average at the end of the present epoch is used to obtain the penalty label. From \cite{huang2017snapshot, laine2016temporal}, because a DNN passes several local minima as training progresses, the prediction confidences are as diverse as those estimated using different networks. This results in a temporal ensemble effect. If we store the prediction confidences per iteration, diverse prediction confidences are obtained, and the penalty label can be estimated stably. Moreover, because the prediction confidences per iteration are already obtained during the forward pass of the training, only they need to be stored. In contrast, single model method requires additional computation to re-predict the training data at the end of each epoch. Therefore, our ensemble method is significantly stable and efficient compared to the single model. Detailed experimental results are presented in Section \ref{Effect_weight_update}.

\subsection{Collaboration with recent methods}
\label{Collaboration}

Criteria\_ALL can be applied to various algorithms for learning with noisy labels. Among them, we apply Criteria\_ALL to Symmetric cross entropy Learning (SL; a recent algorithm in loss correction category) and SELectively reFurbIsh unclEan samples (SELFIE; a recent algorithm in hybrid category).

\subsubsection{SL with Criteria\_ALL}
\label{SL_Criteria_ALL}

To improve Cross Entropy's (CE) learning on hard classes and tolerant to noisy label, SL \cite{wang2019symmetric} introduces $\mathcal{L}_{SL}$ which combines new term, namely Reverse Cross Entropy (RCE), with CE as follows
\begin{eqnarray}
\mathcal{L}_{SL}\left({\mathbf{x}}_{i}, {\hat{\mathbf{y}}}_{i};\boldsymbol{\theta} \right)
=\alpha\mathcal{L}_{CE}\left({\mathbf{x}}_{i}, {\hat{\mathbf{y}}}_{i};\boldsymbol{\theta} \right) + \beta\mathcal{L}_{RCE}\left({\mathbf{x}}_{i}, {\hat{\mathbf{y}}}_{i};\boldsymbol{\theta} \right), \nonumber
\label{eq15}
\end{eqnarray}
where
\begin{eqnarray}
\mathcal{L}_{CE}\left({\mathbf{x}}_{i}, {\hat{\mathbf{y}}}_{i};\boldsymbol{\theta} \right) = -\sum_{j}{\hat{\mathbf{y}}}_{i}\left(j \right)\log P\left(\mathbf{y}=\mathbf{e}^{j}|{\mathbf{x}}_{i};\boldsymbol{\theta} \right),\nonumber
\label{eq16}
\end{eqnarray}
\begin{eqnarray}
\mathcal{L}_{RCE}\left({\mathbf{x}}_{i}, {\hat{\mathbf{y}}}_{i};\boldsymbol{\theta} \right) = -\sum_{j}P\left(\mathbf{y}=\mathbf{e}^{j}|{\mathbf{x}}_{i};\boldsymbol{\theta} \right)\log {\hat{\mathbf{y}}}_{i}\left(j \right),\nonumber
\label{eq17}
\end{eqnarray}
$\alpha$, $\beta$ are the hyperparameters (user design parameters).

In SL, $\mathcal{L}_{SL}$ is applied to entire training data. However, when the sample selection with Criteria\_ALL is used, we can distinguish the clean and noisy samples. Therefore, to increase robustness of DNN for noisy labels, we apply $\mathcal{L}_{SL}$ only to the selected clean samples. In Section \ref{benchmark_datasets}, the effectiveness of SL with Criteria\_ALL is demonstrated.

\subsubsection{SELFIE with Criteria\_ALL}
\label{SELFIE_Criteria_ALL}

\begin{figure}[t]
\centering
\subcaptionbox{\label{fig6:subfig1}}{
    \includegraphics[width=8.3cm]{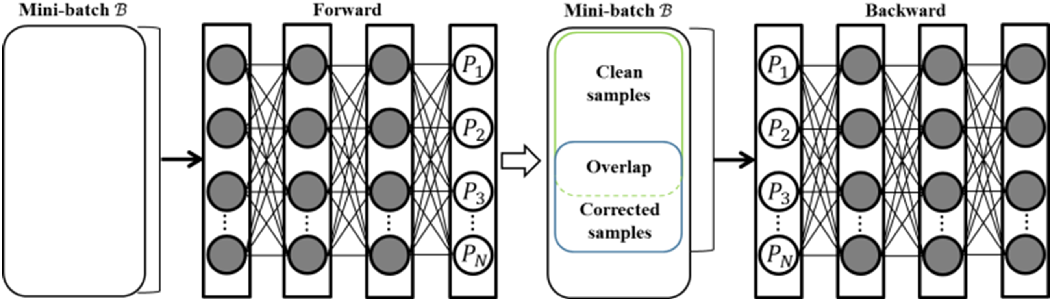}
}\hfill
\subcaptionbox{\label{fig6:subfig2}}{
    \includegraphics[width=8.3cm]{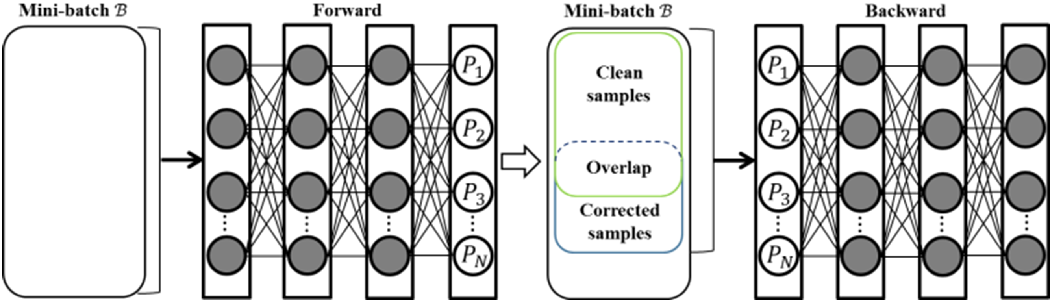}
}
\caption{Training procedure of (a) SELFIE (with Criteria\_OL); (b) SELFIE with Criteria\_ALL; when clean and corrected samples are overlapped, SELFIE trains DNN using samples with corrected label (for corrected samples). Meanwhile, SELFIE with Criteria\_ALL trains DNN using samples with observed label (for clean samples).}
\label{fig6_SELFIE}
\end{figure}

Because hybrid methods are also based on sample selection, Criteria\_ALL can be applied to the SELFIE \cite{song2019selfie} scheme, which achieved the highest performance in its category. In this method, two types of samples (clean and corrected samples) are selected independently at different usage. Clean samples are selected using the conventional sample selection (Criteria\_OL) and corrected samples are selected using the predictive uncertainty that is low if stacked prediction confidences are consistent. Thereafter, DNN is trained using clean samples with the observed label and corrected samples with the estimated label (corrected label) that indicates class of highest prediction confidence. When clean and corrected samples overlap, SELFIE trains DNN using corrected labels (Fig. \ref{fig6_SELFIE}a) because clean samples, selected with Criteria\_OL, have relatively larger error than corrected samples. In addition, to achieve high performance, SELFIE retrains the network several times while preserving the corrected information.

In contrast, clean and noisy samples can be accurately distinguished using SELFIE with Criteria\_ALL compared to using SELFIE with Criteria\_OL. Therefore, we can modify SELFIE to consider the clean samples before the corrected ones when Criteria\_ALL is used (Fig. \ref{fig6_SELFIE}b), i.e., overlapping samples are trained with observed labels, not corrected labels. In Section \ref{benchmark_datasets}, we demonstrate the effectiveness of SELFIE with Criteria\_ALL.

\section{Additional details for experiments}
\subsection{Evaluation metric}
To evaluate the performance of the experiments, we used the following test error and precision definitions: 
\begin{eqnarray}
\textrm{Test error} &=& 1-\frac{\textrm{\# of correct predictions in test data}}{\textrm{\# of test data}}, \nonumber \\
\textrm{Precision} &=& \frac{\textrm{\# of clean samples in selected samples}}{\textrm{\# of selected samples in training data}}. \nonumber 
\end{eqnarray}

\begin{figure*}[t]
\centering
\subcaptionbox{\label{fig9:subfig1} \hspace*{-3.3em}}{
    \includegraphics[width=4cm]{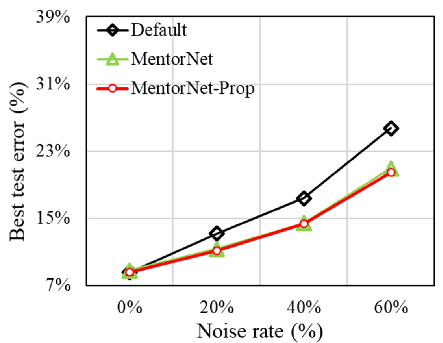}
}
\subcaptionbox{\label{fig9:subfig2} \hspace*{-3.6em}}{
    \includegraphics[width=4cm]{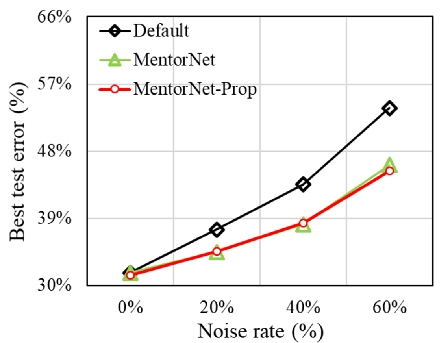}
}
\subcaptionbox{\label{fig9:subfig3} \hspace*{-3.9em}}{
    \includegraphics[width=4cm]{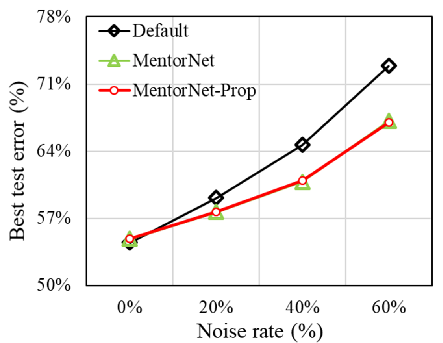}
}
\caption{Best test error (\%) based on various noise rates with symmetry flipping; (a) CIFAR-10; (b) CIFAR-100; (c) Tiny-ImageNet.}
\label{fig9_bench_symmetry}
\end{figure*}

\subsection{Dataset}
\label{Dataset}
We used CIFAR-10 \cite{krizhevsky2009learning}, CIFAR-100 \cite{krizhevsky2009learning}, and Tiny-ImageNet \cite{Tiny_imagenet}, which are widely used benchmark datasets in the noisy label fields \cite{han2018co, song2019selfie, yu2019does, kong2019recycling, arazo2019unsupervised, chen2019understanding, kim2019nlnl}. CIFAR-10 and CIFAR-100 consist of $10$ and $100$ classes, respectively, of $32\,\times\,32$ color images. Each class is composed of subsets with $80$ million categorical images. The numbers of their training and test data are $50,000$ and $10,000$, respectively. Tiny-ImageNet, a subset of ImageNet, consists of $200$ classes. Its numbers of training and test data are $100,000$ and $10,000$, respectively. The experiment was conducted by resizing the images from $64\,\times\,64$ to $32\,\times\,32$. These benchmark datasets consist only of clean samples. Therefore, we artificially corrupted their labels using three commonly used types of noise transition matrices: pair flipping (Fig. \ref{fig7_noise_trainsition_matrix}a), symmetry flipping (Fig. \ref{fig7_noise_trainsition_matrix}b), and mixed flipping (Fig. \ref{fig7_noise_trainsition_matrix}c) \cite{tanaka2018joint, han2018co, song2019selfie, yu2019does, kong2019recycling}. From \cite{han2018co}, pair flipping is a more general case compared to symmetry flipping in the real world. In addition, we conducted experiments based on various noise rates: $\varepsilon = \left \{10\%,20\%,30\%,40\%\right\}$ for pair flipping and $\varepsilon = \left \{0\%,20\%,40\%,60\%\right\}$ for symmetry flipping. Finally, we conducted experiments for mixed flipping with a noise rate 40\% where the labels are corrupted by labels of a class as $\varepsilon_{1}=30\%$ and the other classes as $\varepsilon_{2}=10\%$ (Fig. \ref{fig7_noise_trainsition_matrix}c).

Thereafter, the ANIMAL-10N dataset \cite{song2019selfie} and Clothing1M \cite{xiao2015learning}, whose data are collected from the real world, were employed to examine the performance of Criteria\_ALL. ANIMAL-10N dataset, whose labels were corrupted by human errors, consists of images from 10 animals of similar appearance. The numbers of its training and test data are $50,000$ and $5,000$, respectively, and its noise rate is $8\%$. Furthermore, an experiment was conducted with $64\,\times\,64$ color images without data augmentation or pre-processing. The Clothing1M dataset, which contains 1 million images of clothing, was obtained from online shopping websites with 14 classes. The labels were assigned by surrounding text of images provided by the sellers, and thus the dataset contains severe noisy labels. Its overall accuracy of the labels is $61.54\%$ \cite{xiao2015learning}. The numbers of training and test data were $1,000,000$ and $100,000$, respectively.

\subsection{System setup}
To compare their performances, the experiments were conducted using DenseNet $\left(\mathrm{L} = 25, \mathrm{k} = 12 \right)$ \cite{huang2017densely} with a momentum optimizer $\left(\gamma =0.9 \right)$. Moreover, we have set a batch size of $128$ and applied dropout \cite{srivastava2014dropout} (dropout rate $= 0.2$) and batch normalization \cite{ioffe2015batch}. The learning was conducted for $100$ epochs with an initial learning rate of $0.1$ by reducing the rate by $0.2$ times at the $50$th and $75$th epochs. For the algorithms, the training was conducted without processing for noisy labels until $25$ epochs (warm-up threshold $= 25$) for initial learning, similar to \cite{song2019selfie}. In Co-teaching, we set $\mathrm{T}_{k}=15$ and assumed that $\varepsilon$ is known \cite{han2018co, song2019selfie, wang2019co, yu2019does}. If $\varepsilon$ is unknown in advance, its value can be inferred from the validation set \cite{song2019selfie, liu2015classification, yu2018efficient}. In SL and SL--Prop, we set the hyperparameters as $\alpha = 1$ for all dataset, whereas $\beta$ is set depending on the dataset (CIFAR-10: 0.08, CIFAR-100: 0.3, Tiny-ImageNet: 0.3, ANIMAL-10N: 0.01, Clothing1M: 0.08). In SELFIE, we set the uncertainty threshold to $0.05$, history length to $15$, and restart to $2$ (totally three runs), as in \cite{song2019selfie}. In SELFIE--Prop, samples can be corrected precisely owing to the capability of Criteria\_ALL that distinguishes clean and noisy samples regardless of noise type. Therefore, we set the hyperparameter tightly (uncertainty threshold $= 0.005$ and history length $= 25$) for SELFIE--Prop. Moreover, we set $\lambda$ to 1 by the experiments in Section \ref{hyperparameter} and $\boldsymbol{q}$ is assumed to be uniform in the benchmark and real-world datasets. 
%In extremely noisy label case (symmetry 60\%), we update the weight term in Criteria\_ALL per iteration for stable training.

\subsection{Experimental result on symmetry flipping case}
\label{symmetry_case}
In the symmetry flipping case, because no dominant noisy samples were present, MentorNet--Prop should have exhibited similar performance as MentorNet, as indicated in Theorem \ref{eq14} (Fig. \ref{fig9_bench_symmetry}).

%\begin{table}[t]
%\caption{Best test error (\%) for each restart with noise rate of $40\%$ in CIFAR-100}
%\centering
%\small
%\label{table3}
%\scalebox{0.85}{
%\begin{tabular}{c|cc|cc}
%\hline
%\multirow{2}{*}{Algorithm} & \multicolumn{2}{c|}{Pair flipping}   & \multicolumn{2}{c}{Symmetry flipping} \\ 
 %                                   & SELFIE & \textbf{SELFIE--Prop} & SELFIE  & \textbf{SELFIE--Prop}  \\ \Xhline{3\arrayrulewidth}
%Run 1                      & 46.70           & \textbf{36.58}              & \textbf{38.71}   & 38.72                        \\ 
%Run 2                      & 43.40           & \textbf{35.64}              & \textbf{37.15}   & 37.67                        \\ 
%Run 3                      & 41.30           & \textbf{35.20}              & 37.04            & \textbf{36.38}               \\ \hline
%\end{tabular}}
%\end{table}

\subsection{Experimental result on a real-world dataset}
\label{real_dataset}

\begin{table*}[h]
\centering
\caption{Best test error (\%) in ANIMAL-10N (8\% noise)}
\label{table6}
\resizebox{\textwidth}{!}{%
\begin{tabular}{c|cccccccccc}
\hline
Algorithm                                                       & Default                                              & Active Bias                                          & Co-teaching                                          & Co-teaching+                                         & MentorNet                                            & SL                                                   & SELFIE                                               & \textbf{MentorNet--Prop}                                       & \textbf{SL--Prop}                                              & \textbf{SELFIE--Prop}                                          \\ \Xhline{3\arrayrulewidth}
\setlength{\tabcolsep}{1pt} \renewcommand{\arraystretch}{0.7} 
\begin{tabular}[c]{@{}c@{}}DenseNet\\ (L=25, k=12)\end{tabular} & \setlength{\tabcolsep}{1pt} \renewcommand{\arraystretch}{0.7} \begin{tabular}[c]{@{}c@{}}17.9\\ $\pm$0.02\end{tabular} & \setlength{\tabcolsep}{1pt} \renewcommand{\arraystretch}{0.7} \begin{tabular}[c]{@{}c@{}}17.6\\ $\pm$0.17\end{tabular} & \setlength{\tabcolsep}{1pt} \renewcommand{\arraystretch}{0.7} \begin{tabular}[c]{@{}c@{}}17.5\\ $\pm$0.17\end{tabular} \setlength{\tabcolsep}{1pt} \renewcommand{\arraystretch}{0.7} & \setlength{\tabcolsep}{1pt} \renewcommand{\arraystretch}{0.7} \begin{tabular}[c]{@{}c@{}}23.6\\ $\pm$0.20\end{tabular} & \setlength{\tabcolsep}{1pt} \renewcommand{\arraystretch}{0.7} \begin{tabular}[c]{@{}c@{}}18.0\\ $\pm$0.15\end{tabular} & \setlength{\tabcolsep}{1pt} \renewcommand{\arraystretch}{0.7} \begin{tabular}[c]{@{}c@{}}17.3\\ $\pm$0.05\end{tabular} \setlength{\tabcolsep}{1pt} \renewcommand{\arraystretch}{0.7} & \setlength{\tabcolsep}{1pt} \renewcommand{\arraystretch}{0.7} \begin{tabular}[c]{@{}c@{}}17.0\\ $\pm$0.10\end{tabular} & \setlength{\tabcolsep}{1pt} \renewcommand{\arraystretch}{0.7} \begin{tabular}[c]{@{}c@{}}16.5\\ $\pm$0.19\end{tabular} & \setlength{\tabcolsep}{1pt} \renewcommand{\arraystretch}{0.7} \begin{tabular}[c]{@{}c@{}}16.3\\ $\pm$0.04\end{tabular} \setlength{\tabcolsep}{1pt} \renewcommand{\arraystretch}{0.7} & \setlength{\tabcolsep}{1pt} \renewcommand{\arraystretch}{0.7} \begin{tabular}[c]{@{}c@{}}\textbf{16.1}\\ \textbf{$\pm$0.13}\end{tabular} \\ \hline
\end{tabular}%
}
\end{table*}

We applied the algorithms on the ANIMAL-10N and Clothing1M to examine the effectiveness of Criteria\_ALL in real-world datasets. From \cite{song2019selfie}, the ANIMAL-10N dataset was influenced by human errors during the image labeling process, because the classes consist of five pairs of ``confusing'' animals. The noise rate ($\varepsilon=8\%$) was estimated by cross-validation with grid search \cite{song2019selfie}. The experimental results indicate that the test error of MentorNet--Prop was $0.5\%$ lower than that of SELFIE, which exhibited the lowest test error among the conventional algorithms (Table \ref{table6}). Furthermore, SL--Prop and SELFIE--Prop achieved a test error $1.0\%$ and $0.9\%$ lower than that of SL and SELFIE, respectively. Because the noise rate was small ($8\%$), the difference in performance was small. However, the experimental results indicate that Criteria\_ALL is still effective.

\begin{table*}[!h]
\centering
\caption{Test error (\%) in Clothing1M}
\label{table7}
\resizebox{\textwidth}{!}{%
\begin{tabular}{c|cccccccccc}
\hline
Algorithm            & Default   & \begin{tabular}[c]{@{}c@{}}VAT\\ \cite{Miyato2018virtual}\end{tabular} & \begin{tabular}[c]{@{}c@{}}F--correction\\ \cite{patrini2017making}\end{tabular} & \begin{tabular}[c]{@{}c@{}}GCE\\ \cite{Zhang2018generalized}\end{tabular} & \begin{tabular}[c]{@{}c@{}}LCCN\\ \cite{Yao2019safeguarded}\end{tabular} & \begin{tabular}[c]{@{}c@{}}DMI\\ \cite{xu2019l_dmi}\end{tabular} & MentorNet                                            & SL                                                   & \textbf{MentorNet-Prop}                                       & \textbf{SL-Prop}                                           \\ \Xhline{3\arrayrulewidth}
ResNet-50 & 31.06   & 30.43                                                  & 29.17  & 30.91                                                    & 28.37                                                    & 27.54 & 30.70     & 29.40 & 27.44          & \textbf{26.98}   \\ \hline
\end{tabular}
}
\end{table*}

Experiments for Clothing1M were conducted using pretrained ResNet-50 as \cite{xu2019l_dmi}. We fine-tuned the network for 10 epochs where the learning rate was $\num{1.0e-6}$ for the first 5 epochs and $\num{0.5e-6}$ in the next 5 epochs. Several methods in loss correction \cite{patrini2017making, xu2019l_dmi, Miyato2018virtual, Zhang2018generalized, Yao2019safeguarded} were compared with the proposed method, and we used the results reported by \cite{xu2019l_dmi}. Experimental results indicate that our MentorNet--Prop had lower test errors than whole conventional algorithms (Table \ref{table7}). When the loss of MentorNet is changed to the loss of SL (SL--Prop), we can achieve $0.46\%$ lower test error than MentorNet--Prop. Consequently, we can show that the Criteria\_ALL is effective for real-world datasets.

\subsection{Effect of update method for penalty label}
\label{Effect_weight_update}

It is important to estimate a penalty label accurately when Criteria\_ALL is used. To increase the accuracy of the penalty label, we propose the temporal ensemble method (Section \ref{subsub_update}; Algorithm \ref{algo3}). With regard to the update method, Algorithm \ref{algo3} exhibited a lower test error than that of Algorithm \ref{algo2} for pair and symmetry flipping (Table \ref{table5}). Because the predictions were stored per iteration, diverse predictions were obtained, and thus the penalty label can be estimated stably using the ensemble effect. Furthermore, the difference between these two update methods was apparent for pair flipping because penalty label is significantly effective for that noise type.

\begin{table}[h]
\caption{Best test error (\%) of MentorNet--Prop depending on update methods for penalty label and considering noise rate of $40\%$}
\centering
\small
\label{table5}
\scalebox{0.88}{
\begin{tabular}{c|c|c|c}
\hline
Dataset                        & Method & Pair flipping & Symmetry flipping \\ \Xhline{3\arrayrulewidth}
\multirow{2}{*}{CIFAR-10}      & Algorithm 2    & 30.8          & 14.5 \\    

                               & \textbf{Algorithm 3}      & \textbf{12.2}          & \textbf{14.4} \\\hline
                               
\multirow{2}{*}{CIFAR-100}     & Algorithm 2    & 42.9          & 39.4\\
                               & \textbf{Algorithm 3}      & \textbf{36.6}          & \textbf{38.4} \\ \hline
                               
\multirow{2}{*}{Tiny-ImageNet} & Algorithm 2    & 73.0          & 62.1               \\
                               & \textbf{Algorithm 3}      & \textbf{62.5}          & \textbf{61.0}\\ \hline
                               
\end{tabular}}

\end{table}

\subsection{Experiments on the hyperparameter}
\label{hyperparameter}

\begin{figure}[h]
\centering
\includegraphics[width=6cm]{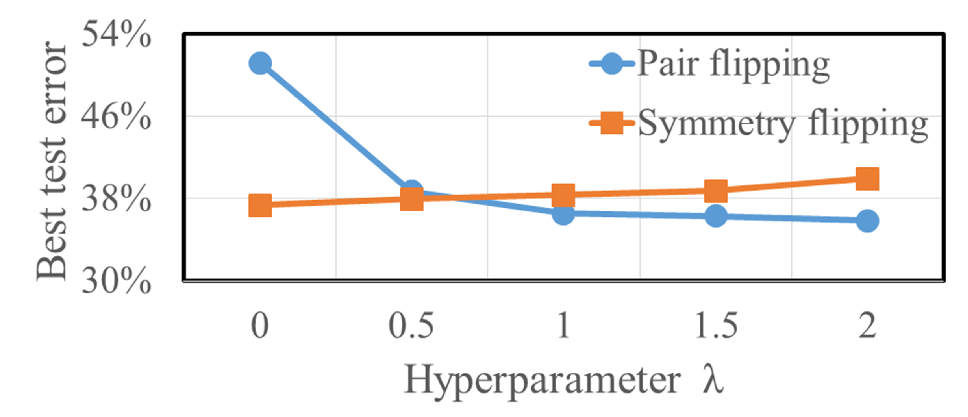}
\caption{Experiments on hyperparameter at noise rate $40\%$ in CIFAR-100.}
\label{fig10_hyperparameter}
\end{figure}

The Criteria\_ALL exhibits only one hyperparameter $\lambda$ that adjusts the influence between Criteria\_OL and Criteria\_PL. To analyze the effect of $\lambda$, experiments were conducted with several values of $\lambda\,\left(=\left \{0.0,0.5,1.0,1.5,2.0\right \}\right)$, at a noise rate of $40\%$ in CIFAR-100 for both pair and symmetry flipping (Fig. \ref{fig10_hyperparameter}). In pair flipping, the test error decreased steeply when $\lambda$ was lower than one owing to the high effectiveness of the penalty label. In symmetry flipping, the test error slightly increased with increasing $\lambda$, owing to the incorrect penalty label. Therefore, we empirically set $\lambda=1$ to consider the trade-off between pair and symmetry flipping.

\subsection{Discussion}
\label{Discussion}
In this section, we discuss the effect of Criteria\_OL, Criteria\_PL, and Criteria\_ALL on sample selection, and the difference between penalty label and noise transition matrix for pair and symmetry flipping cases.

\subsubsection{Comparison of precisions of the criteria}
\begin{figure}[h]
\centering
\subcaptionbox{\label{fig10:subfig1}\hspace*{-3.0em}}{
    \includegraphics[width=3.9cm]{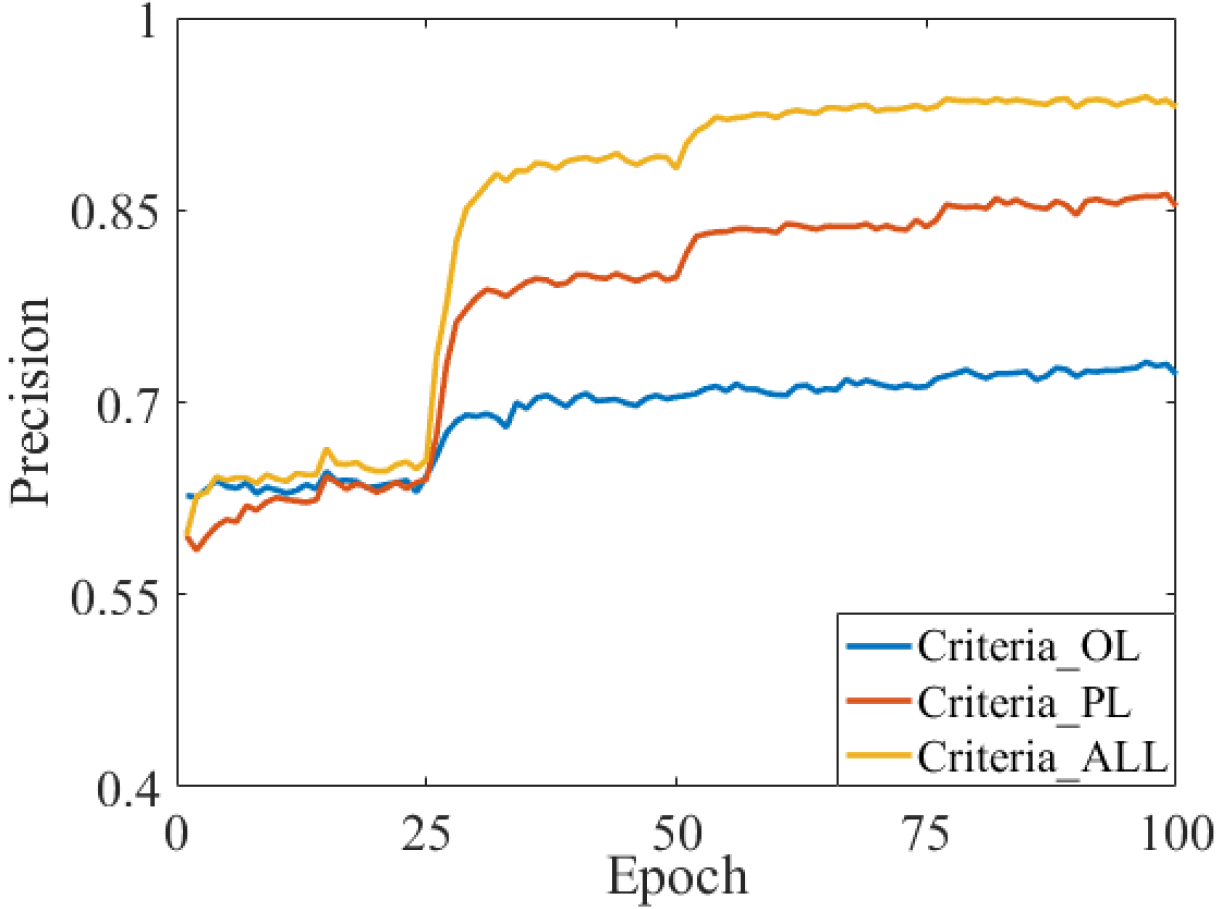}
}
\subcaptionbox{\label{fig10:subfig2}\hspace*{-3.1em}}{
    \includegraphics[width=3.9cm]{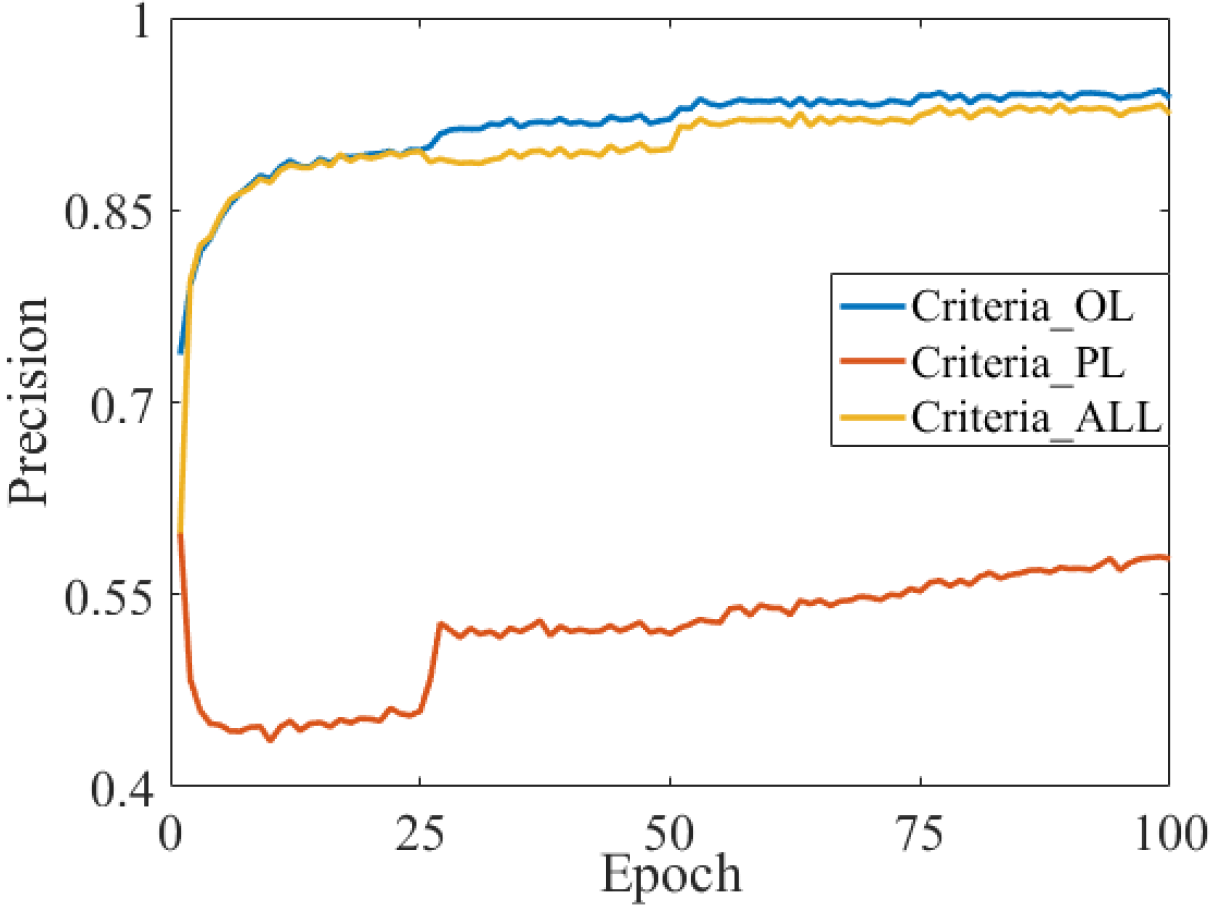}
}
\caption{Precision of Criteria\_OL, Criteria\_PL, and Criteria\_ALL at noise rate $40\%$ in CIFAR-100; (a) pair flipping; (b) symmetry flipping.}
\label{fig11_precision}
\end{figure}

The precisions of Criteria\_OL, Criteria\_PL, and Criteria\_ALL were compared to examine the performance in selecting clean samples for each epoch (Fig. \ref{fig11_precision}). Each criteria was applied to MentorNet from the $25$th epoch (warm-up). In addition, the learning rate was reduced by $0.2$ times at the $50$th and $75$th epochs. In the pair flipping case, owing to the shared patterns of dominant noisy samples, severe memorization occurred. This results in a precision of approximately $70\%$ for Criteria\_OL. In contrast, owing to the penalty applied to the dominant noisy samples, the precision of Criteria\_PL was approximately $15\%$ higher than that of Criteria\_OL. Furthermore, Criteria\_ALL, which uses the two complementary criteria simultaneously, exhibited the highest precision (approximately $92\%$), because Criteria\_OL and Criteria\_PL evaluate samples through different patterns caused by observed label and penalty label. 

In symmetry flipping, because there were no dominant noisy samples, Criteria\_OL exhibited a high precision (approximately $92\%$), whereas Criteria\_PL did not affect the selection of clean samples (approximately $60\%$ precision). Initially, Criteria\_ALL exhibited lower precision than Criteria\_OL. However, as the training stage progressed, the accuracy of estimation of the penalty labels increased steadily, and the precision was similar to that of Criteria\_OL.

\subsubsection{Comparison between Criteria\_ALL and noise transition matrix}
\label{Comparison noise transition matrix}
\begin{table}[t]
\caption{Best test error (\%) of Default, loss correction with noise transition matrix (F--correction), and MentorNet--Prop with noise rate of $40\%$ in Tiny-ImageNet}
\centering
\small
\label{table8}
\scalebox{0.9}{
\begin{tabular}{c|ccc}
\hline
Algorithm         & Default & F--correction & \textbf{MentorNet--Prop} \\ \Xhline{3\arrayrulewidth}
Pair flipping     & 69.9             & 72.8                  & \textbf{62.5}                  \\ 
Symmetry flipping & 64.7             & 75.3                  & \textbf{61.0}                  \\ \hline
\end{tabular}}
\end{table}

The average-prediction confidence of each observed label (which is used to estimate the penalty label) can also be used to estimate the noise transition matrix. Therefore, the performance of the sample selection using Criteria\_ALL (MentorNet--Prop) was compared with that of F--correction \cite{patrini2017making}, a representative method of loss correction with a noise transition matrix. In this case, the noise transition matrix was replaced by the average-prediction confidence used to estimate proposed penalty label in Criteria\_PL. From \cite{han2018co, song2019selfie,jiang2017mentornet}, it is challenging to estimate the noise transition matrix when the number of classes is large. Consequently, the noise transition matrix, estimated by the average-prediction confidence of each observed label in Tiny-ImageNet ($200$ classes), was inaccurate. Furthermore, F--correction does not distinguish clean and noisy samples, unlike MentorNet--Prop. Therefore, in Table \ref{table8}, F--correction achieved a test error even $2.9$--$10.6\%$ higher than that of the Default. In contrast, because in Criteria\_ALL, only relative average-prediction confidence is needed to significantly penalize the dominant noisy samples unlike noise transition matrix, MentorNet--Prop exhibited remarkable performance regardless of the number of classes (Tables \ref{table2}, \ref{table8}).

%\section{Acknowledgments}
%This research was supported in part by the MSIT(Ministry of Science and ICT), Korea, under the ICT Consilience Creative program(IITP-2019-2011-1-00783) supervised by the IITP(Institute for Information \& communications Technology Planning \& Evaluation), National Research Foundation of Korea (NRF) funded by the Korean government (2020R1C1C1011857, 2020M3C1B8081320), Basic Science Research Program through the National Research Foundation of Korea (NRF) funded by the Ministry of Education (2020R1A6A3A01098940), and LG Display under LGD-POSTECH Research Program.

%\bibliography{example_paper}
%\bibliographystyle{icml2021}





%\bibliography{example_paper}
%\bibliographystyle{icml2021}

\newpage
\onecolumn
\section{Special case: Symmetry flipping}
\label{Special cases Symmetry}
If the label of a true class is corrupted by the labels of the other classes with the same noise rate (symmetry flipping), Criteria\_PL should not affect the selection of clean samples because there are no dominant noisy samples and penalty label has same value. This property is stated in the following Theorem.

\newtheorem{thm}{Theorem}
\begin{thm}
\label{eq14}
Supposing that the noise type is symmetry flipping with ideal penalty label  ($\mathit{ \textit{\textbf{m}}^{k}=\frac{1}{K-1}}$), the order of samples with Criteria\_ALL is identical to that with Criteria\_OL as
\begin{eqnarray}
{\mathrm{ORDER}}_{\left({\mathbf{x}}_{i},{\hat{\mathbf{y}}}_{i} \right)\in {\mathcal{B}}}^{\mathrm{Descent}} \mathrm{Criteria\_ALL}\left({\mathbf{x}}_{i},{\hat{\mathbf{y}}}_{i}, {\tilde{\mathbf{y}}}_{i};{\boldsymbol{\theta}} \right) = {\mathrm{ORDER}}_{\left({\mathbf{x}}_{i},{\hat{\mathbf{y}}}_{i} \right)\in {\mathcal{B}}}^{\mathrm{Descent}} \mathrm{Criteria\_OL}\left({\mathbf{x}}_{i},{\hat{\mathbf{y}}}_{i};{\boldsymbol{\theta}} \right). \nonumber
\end{eqnarray}
\end{thm}

Proof.
\begin{eqnarray}
\mathrm{Criteria\_ALL}\left({\mathbf{x}}_{i},{\hat{\mathbf{y}}}_{i}, {\tilde{\mathbf{y}}}_{i};{\boldsymbol{\theta}} \right) \nonumber
&=& \sum_{j}{\hat{\mathbf{y}}}_{i}\left(j \right)P\left(\mathbf{y}=\mathbf{e}^{j}|{\mathbf{x}}_{i};\boldsymbol{\theta} \right) - \lambda\sum_{j}{\tilde{\mathbf{y}}}_{i}\left(j \right)P\left(\mathbf{y}=\mathbf{e}^{j}|{\mathbf{x}}_{i};\boldsymbol{\theta} \right) \nonumber \\
&=& \sum_{j}{\mathbf{e}^{k}}\left(j \right)P\left(\mathbf{y}=\mathbf{e}^{j}|{\mathbf{x}}_{i};\boldsymbol{\theta} \right) - \lambda\sum_{j}{\mathbf{m}^{k}}\left(j \right)P\left(\mathbf{y}=\mathbf{e}^{j}|{\mathbf{x}}_{i};\boldsymbol{\theta} \right) \quad \left(\because {\hat{\mathbf{y}}}_{i} = {\mathbf{e}^{k}}, {\tilde{\mathbf{y}}}_{i} = {\mathbf{m}^{k}} \right) \nonumber \\
&=& \sum_{j=k}P\left(\mathbf{y}=\mathbf{e}^{j}|{\mathbf{x}}_{i};\boldsymbol{\theta} \right) - \frac{\lambda}{K-1}\sum_{j \neq k}P\left(\mathbf{y}=\mathbf{e}^{j}|{\mathbf{x}}_{i};\boldsymbol{\theta} \right) \quad \nonumber \\ &&\left(\because \mathrm{In} \ \mathrm{ideal} \ \mathrm{case,} \ {\mathbf{m}^{k}}\left(j \right) = \frac{1}{K-1} \, \mathrm{for} \; j \neq k\right) \nonumber \\
&=& \sum_{j=k}P\left(\mathbf{y}=\mathbf{e}^{j}|{\mathbf{x}}_{i};\boldsymbol{\theta} \right) - \frac{\lambda}{K-1}\left(1-\sum_{j=k}P\left(\mathbf{y}=\mathbf{e}^{j}|{\mathbf{x}}_{i};\boldsymbol{\theta} \right) \right) \nonumber\\ &&\left(\because \sum_{j}P\left(\mathbf{y}=\mathbf{e}^{j}|{\mathbf{x}}_{i};\boldsymbol{\theta} \right) = 1\right) \nonumber \\
&=& \left(1+\frac{\lambda}{K-1} \right)\sum_{j=k}P\left(\mathbf{y}=\mathbf{e}^{j}|{\mathbf{x}}_{i};\boldsymbol{\theta} \right) - \frac{\lambda}{K-1} \nonumber \\
&=& \left(1+\frac{\lambda}{K-1} \right)\sum_{j}{\hat{\mathbf{y}}}_{i}\left(j \right)P\left(\mathbf{y}=\mathbf{e}^{j}|{\mathbf{x}}_{i};\boldsymbol{\theta} \right) - \frac{\lambda}{K-1} \nonumber \\
&=& \left(1+\frac{\lambda}{K-1} \right)\mathrm{Criteria\_OL}\left({\mathbf{x}}_{i},{\hat{\mathbf{y}}}_{i};{\boldsymbol{\theta}} \right) - \frac{\lambda}{K-1}.  \nonumber
\end{eqnarray}

Because the scaling and bias do not alter the order, the order of samples with both the criteria is identical.                                  \quad $\square$ 

%In Fig. \ref{fig3:subfig3} where only few non-dominant noisy samples are included in the selected samples, the number of the selected non-dominant noisy samples of Criteria\_OL and Criteria\_ALL are equal although the scaling and bias differ.

\end{document}